\documentclass[lettersize,journal]{IEEEtran}
\usepackage{amsmath,amsfonts}
\usepackage{algorithmic}
\usepackage{algorithm}
\usepackage{array}
\usepackage[caption=false,font=normalsize,labelfont=sf,textfont=sf]{subfig}
\usepackage{textcomp}
\usepackage{stfloats}
\usepackage{url}
\usepackage{verbatim}
\usepackage{graphicx}
\usepackage{epstopdf}
\usepackage{cite}
\usepackage{makecell}
\usepackage{titlesec}
\begin{document}

\title{Learning Surface Scattering Parameters From SAR Images Using Differentiable Ray Tracing}

\author{Jiangtao Wei, ~\IEEEmembership{Student Member, IEEE}, Yixiang Luomei, ~\IEEEmembership{Member, IEEE}, Xu Zhang, ~\IEEEmembership{Student Member, IEEE}, and Feng Xu, ~\IEEEmembership{Senior Member, IEEE}

\thanks{The authors are with the Key Laboratory for Information Science of
Electromagnetic Waves (MoE), Fudan University, Shanghai 200433, China
(Corresponding author: Feng Xu, e-mail: fengxu@fudan.edu.cn).}}

\markboth{submitted}
{Shell \MakeLowercase{\textit{et al.}}: A Sample Article Using IEEEtran.cls for IEEE Journals}


\maketitle

\begin{abstract}
Simulating high-resolution Synthetic Aperture Radar (SAR) images in complex scenes has consistently presented a significant research challenge. The development of a microwave-domain surface scattering model and its reversibility are poised to play a pivotal role in enhancing the authenticity of SAR image simulations and facilitating the reconstruction of target parameters. Drawing inspiration from the field of computer graphics, this paper proposes a surface microwave rendering model that comprehensively considers both Specular and Diffuse contributions. The model is analytically represented by the coherent spatially varying bidirectional scattering distribution function (CSVBSDF) based on the Kirchhoff approximation (KA) and the perturbation method (SPM). And SAR imaging is achieved through the synergistic combination of ray tracing and fast mapping projection techniques. Furthermore, a differentiable ray tracing (DRT) engine based on SAR images was constructed for CSVBSDF surface scattering parameter learning. Within this SAR image simulation engine, the use of differentiable reverse ray tracing enables the rapid estimation of parameter gradients from SAR images. The effectiveness of this approach has been validated through simulations and comparisons with real SAR images. By learning the surface scattering parameters, substantial enhancements in SAR image simulation performance under various observation conditions have been demonstrated.
\end{abstract}


\section{Introduction}
\IEEEPARstart{S}{ynthetic} aperture radar (SAR) has emerged as a critical remote sensing technology, owing to its ability to provide all-weather, long-range, and penetrating observations. Advances in technology enable SAR systems to acquire high-resolution remote sensing images of large-scale, complex scenes across spaceborne, airborne and other platforms, providing rich target information. This development subsequently created the need for SAR image interpretation, where corresponding forward problem SAR image simulation is fundamental research for solving the inverse problem.

However, the main limitation of current SAR image simulation lies in its focus on computations for a single target or small-scale scenes, with substantial challenges remaining in simulating large-scale scenarios. In detail, this issue can be attributed to several challenges: firstly, the difficulty in acquiring accurate geometric and material information of targets; secondly, existing simulation engines generally rely on numerical discretization methods, resulting in exceedingly high computational complexity; and finally, the inherent complexity of electromagnetic scattering mechanism, further complicating the simulation process.

Currently, the main research methodologies for SAR image simulation include conventional computational electromagnetics and imaging methods, data-based approaches, and the recently emerged differentiable SAR renderers. Details are as follows:
\subsection{Traditional computational electromagnetics and imaging methods}
Conventional methods are based on scattering theory and the principles of SAR imaging. The scattering models are used to calculate the scattering results of random rough surface and targets, which mainly include numerical methods, analytical methods, and their hybrid approaches. The computational complexity of numerical methods such as the method of moment (MoM) \cite{harrington1996field}, finite difference time-domain (FDTD) \cite{yee1966numerical}, time-dmain integral equation (TDIE) \cite{weile2004novel}, finite element method (FEM) \cite{jin2015finite} increases with the size of the target, making them unsuitable for large-scale scene simulation calculations. Therefore, this article mainly reviews analytical methods. A variety of analytical scattering models are proposed for random rough surface. The corresponding scattering models include Kirchhoff approximation (KA) \cite{beckmann1987scattering}, small perturbation method (SPM) \cite{wang1986electromagnetic}, double-scale model and integral equation method (IEM) \cite{fung1992backscattering}. These methodologies primarily focus on solving the Muller matrix of rough surfaces.


The most commonly used method to calculate the scattering of electrically large targets is the high-frequency approximation method. It includes current-based physical optics (PO) \cite{kouyoumjian1965asymptotic}, physical theory of diffraction (PTD)\cite{ufimtsev1991elementary} and ray-based methods such as geometrical optics (GO), geometrical theory of diffraction (GTD)\cite{keller1962geometrical}, uniform geometric theory of diffraction (UTD)\cite{kouyoumjian1974uniform}, ray tracing (RT)\cite{deschamps1972ray}, and shooting and bouncing ray (SBR)\cite{ling1989shooting}, etc.
The physical optics method involves calculating the induced current on the target's surface, followed by an integration of this surface current to determine the scattered field. This approach ensures a spatial distribution of the scattering field devoid of singular points. However, its limitation lies in the inability to adequately address multiple electromagnetic wave scattering. On the other hand, ray-based methods treat electromagnetic waves as rays, utilizing ray tracing to ascertain each ray's contribution. These methods demonstrate proficiency in handling multiple scattering phenomena, particularly with complex targets.

In addition, the bidirectional reflection distribution function (BRDF) is also an analytical model to characterize the electromagnetic scattering characteristics of the environment and targets. For example, \cite{norman1985contrasts} uses a simple model to predict soil and canopy BRDF, \cite{tomiyasu1988relationship} studies the relationship between the scattering coefficient and BRDF of rough surfaces. \cite{woodham1987analytic} studies radiation correction and parameter inversion, \cite{chang2020assessment} studies surface albedo, and \cite{liang1993analytic} studies canopy biophysical properties. In recent years, Zhang and Xu\cite{zhang2021coherent} propose the coherent spatially varying bidirectional scattering distribution function (CSVBSDF) of rough surfaces. However, this model does not consider the coupling between SAR image pixels and does not provide a method to learn surface physical properties from measured SAR images.

\subsection{Data-based methods}
The data-based SAR images generation methods uses a deep generative model \cite{goodfellow2020generative} to directly generate SAR images. This type of method generate samples similar to training samples through the game between the generative model and the discriminative model. Some researchers tried to generate SAR images using Generative Adversarial Network (GAN) \cite{ guo2017synthetic }, Wasserstein GAN \cite{cui2019image} and causal adversarial autoencoders for disentangled SAR image representation \cite{ guo2023interpretable }. The role of generated samples is explored to extend the training images for SAR target recognition.

However, SAR target images are attitude-sensitive, which means that SAR images of the same target at different azimuth angles will be very different. This type of method relies heavily on existing SAR images, lacks scattering mechanisms, and is difficult to interpret. Currently, they are mostly used for expanded samples in target recognition, which poses great challenges to the generation of SAR data without corresponding observations.

\subsection{Differential simulation engine}
Differentiable rendering is a simulation engine that encompasses both forward and inverse processes, capable of differentiation and derivation. By ensuring the differentiability of the image generation process, it enhances image analysis and processing capabilities through modern optimization techniques, thereby improving system performance and efficiency across various applications. This is crucial for tasks such as parameter estimation, image reconstruction, or model training using optimization algorithms like gradient descent. Differentiable rendering evolves from traditional rendering models, integrating innovative differentiable techniques. In optics, differentiable rendering techniques have made substantial progress, with notable examples including OpenDR \cite{loper2014opendr}, SoftRas \cite{liu2019soft}, DIRT \cite{li2018differentiable}, DIBR \cite{henderson2020learning}, Redner \cite{chen2019learning}, PSDR \cite{zhang2022path}, and more. 

In terms of SAR, there exists several simulation engines, such as bidirectional analytical ray tracing (BART) \cite{xu2009bidirectional}, PolSARpro \cite{pottier2009overview} and RaySAR\cite{auer20113d} etc. They proficient in computing the target's backscattered field. The mapping and projection algorithm (MPA) \cite{xu2006imaging} stands out in enhancing simulation efficiency within complex scenes, tackling vegetation, buildings, and rough terrain while concentrating on the intricacies of scattering, extinction, mapping, and projection processes. Nevertheless, these simulation engines, when applied to complex targets or scenes, necessitate a certain degree of prior knowledge and parameter setting, including the physical properties and geometric structure of the target or scene. Acquiring this authentic prior knowledge is a formidable challenge, and manual parameter setting is not only time-consuming but also prone to inaccuracies.

The widespread deployment of radar platforms has democratized access to SAR images of numerous scenes. Drawing inspiration from differentiable rendering techniques in optics, developing differentiable SAR renderers is a promising avenue. This renderer could synergize parameter inversion based on existing SAR images with established scattering mechanism models. Starting with the inversion of the target's physical properties and geometry from measured SAR images, it proceeds to forward simulation, culminating in a SAR image simulation method tailor-made for large-scale scenes. This methodology not only augments the efficacy of SAR image simulation in extensive scenes but also enhances the interpretability of SAR images. The Differentiable SAR Renderer (DSR) proposed by our previous work \cite{fu2022differentiable} exemplifies such a renderer, capable of forward rendering from 3D geometric structures to SAR images and reverse reconstruction from SAR images back to 3D geometric structures. However, the rasterization approximation method of DSR limits its applicability. It mainly solves single scattering and geometric reconstruction, and it is difficult to cope with the challenge of multiple scattering. Consequently, there is a need to design a method that can learn spatially varying surface scattering parameters from SAR images for more comprehensive simulation and analysis.

To this end, this paper aims to develop a surface scattering parameter learning method for CSVBSDF based on microwave scattering model. A differentiable ray tracing (DRT) simulator for SAR imaging has been implemented. The accuracy and effectiveness of the method are verified through simulation and measured comparisons. The contributions of this work are summarized as follows:

1)	A surface microwave rendering model is proposed that considers the specular and diffuse reflection contributions of microfacets based on a combination of KA and SPM. The single-ray scattering intensity is calculated through the surface microwave rendering model and Monte Carlo sampling is performed to improve accuracy, and SAR is performed through fast mapping projection imaging.

2)	A novel differentiable ray-tracing SAR simulator DRT is implemented, which guarantees that the mapped projection SAR imaging and surface microwave rendering models are differentiable, and enables learn spatially varying surface scattering parameters based on the microwave rendering model, and unbiased estimation of material parameters for complex scenes with only a small number of observation viewpoints.

3)	Extensive validation and estimation of the method have been conducted. Inverse learning of surface scattering parameters is performed based on simulated and measured SAR images, and similarity validation is performed in other observation conditions. Discuss the effect of surfels visibility on the learning results, and ensure that a surface element is observed at least once to determine the scattering results of that surface element in the corresponding observation viewpoint.

The remainder of this article is organized as follows. Section II introduces the DRT forward model and derives an analytical form of the SAR image as a function of target geometry and radar configuration. Then the inverse reconstruction framework is introduced in Section III, where the loss function is established and gradient estimation is performed. The section IV is the experiment, which conducts simulation and differentiable learning on complex targets, and conducts surface scattering parameter learning through simulated SAR images and measured SAR images to verify the effectiveness of DRT. The time consumption of forward simulation and backward inverse learning is also discussed. Section V concludes the paper.

\section{Forward Model}
\begin{figure}[!t]
\centering
\includegraphics[width=3.5in]{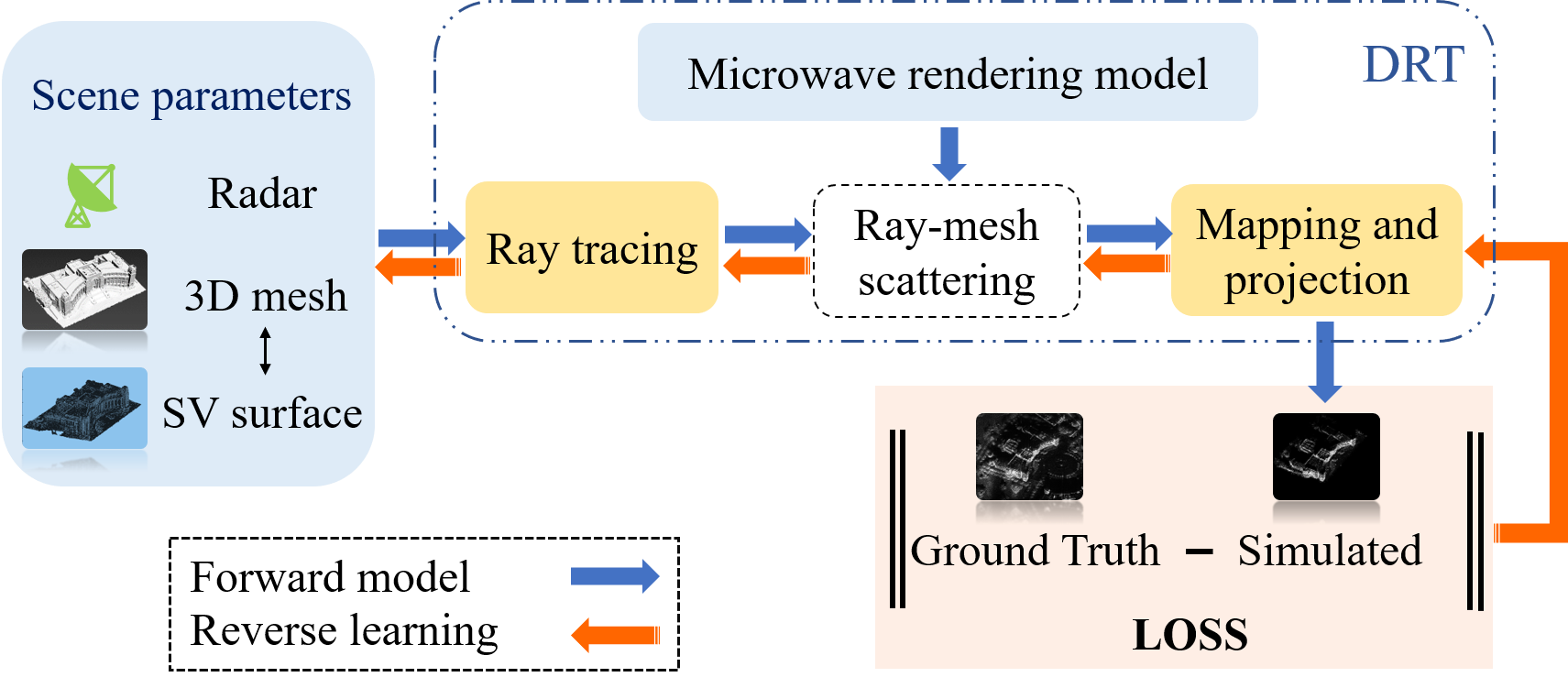}
\caption{SAR forward model and backward differentiable learning framework.}
\label{fig_1}
\end{figure}

A differentiable ray tracing SAR simulator is proposed in this paper, called DRT. Fig. \ref{fig_1} shows the overall computing framework of the forward model and backward differentiable learning. The radar parameters describe the observation conditions, the 3D grid geometry of the scene and the spatially varying (SV) surface describe the specific scattering parameters of the target scene. The relationship between scattering intensity and scene parameters is constructed through a surface microwave rendering model. This section mainly introduces the forward SAR image simulation model based on ray tracing.

Regarding the forward model, multiple rays containing electromagnetic wave information are emitted from the radar according to the radar observation parameters. The ray hits the triangular surface element in the target geometry. The geometric information of the hit point, the surface scattering properties and the scattering intensity of each ray are calculated based on ray tracing. Finally, the SAR scattering intensity image at the observation viewpoint is obtained through mapping and projection. Therefore, the relationship between the simulated SAR image and each parameter can be constructed.

\subsection{Optical reflection model}
There are various rendering models that have been proposed in the field of optics, such as Phong BRDF \cite{phong1998illumination}, microfacet BRDF \cite{dong2015predicting}, wave optics-based BRDFs \cite{yan2018rendering} and Disney BRDF \cite{karis2013real}, etc. In recent years, rendering based on the Disney BRDF model has yielded very realistic results, which comparisons almost identical to actual camera observations. The Disney BRDF model includes diffuse BRDF items and specular BRDF items. A general form of the microfacet model for isotropic materials is:

\begin{equation}
\label{deqn_ex1}
f\left( {\boldsymbol{k}_{i}},{\boldsymbol{k}_{o}} \right)={{f}_\text{diffuse}}+{{f}_\text{specular}}.
\end{equation}

The diffuse BRDF term uses Lambertian diffuse reflection model:

\begin{equation}
\label{deqn_ex2}
{f}_\text{diffuse}=\frac{\rho }{\pi }.
\end{equation}

\noindent where \(\rho\) is the diffuse albedo of the material. Specular BRDF is represented as follows:

\begin{equation}
\label{deqn_ex3}
{f}_\text{specular}({\boldsymbol{k}_{i}},{\boldsymbol{k}_{o}},\boldsymbol{n})=\frac{D({\boldsymbol{k}_{h}})F({\boldsymbol{k}_{i}},{\boldsymbol{k}_{h}})G({\boldsymbol{k}_{i}},{\boldsymbol{k}_{h}},{\boldsymbol{k}_{o}})}{4(\boldsymbol{n}\cdot {\boldsymbol{k}_{i}})(\boldsymbol{n}\cdot {\boldsymbol{k}_{o}})}.
\end{equation}

\noindent where ${\boldsymbol{k}_{i}}$ and ${\boldsymbol{k}_{o}}$ represent the incident and outgoing directions; $\boldsymbol{n}$ is the normal of microfacets; ${\boldsymbol{k}_{h}}={({\boldsymbol{k}_{i}}+{\boldsymbol{k}_{o}})}/{\left\| {\boldsymbol{k}_{i}}+{\boldsymbol{k}_{o}} \right\|}$ is the half-angle vector between ${\boldsymbol{k}_{i}}$ and ${\boldsymbol{k}_{o}}$, $f$ is the BRDF; $D$ is the normal distribution function (NDF); $G$ is the shadowing-masking term; $F$ is Fresnel term. Inspired by Disney BRDF model, a microfacet scattering rendering model be constructed in microwave domain for SAR imaging simulation.

\subsection{Surface microwave rendering model}
\subsubsection{Surface microwave rendering model}

In the traditional rough surface scattering model, a scattering coefficient is obtained by integrating over the entire rough surface assuming that the rough surface is infinite, e.g., ground, sea surface, etc. It also assumes that the entire rough surface can be viewed as an ergodic stochastic process defined by power spectral density (PSD) function. The scattering coefficient $\sigma$ in the microwave domain is defined as the energy ratio between the scattered field, which uniformly fills the entire solid angle, and the incident field.:

\begin{equation}
\label{deqn_ex4}
{{\sigma }_{pq}}=4\pi H_{pq}^{2}.
\end{equation}

\noindent where ${{H}_{pq}}$ represents each element of the coherent scattering complex matrix with $p$ polarization incident and $q$ polarization scattering. When calculating the random rough surface scattering field, KA and SPM corresponds to the contribution of Specular and Diffuse, respectively. Therefore, this article attempts to propose a surface microwave rendering model based on KA and SPM. Coefficient $0 \leq \tau \leq 1$ is used to adjust the proportion of specular and diffuse contributions. The microwave scattering model of surface is expressed as:

\begin{equation}
\label{deqn_ex5}
\mathcal{S}(\boldsymbol{p},r,\theta ,\boldsymbol{\zeta} ) = {{\sigma }^\text{(SPM+KA)}}=(1-\tau ){{\sigma }^\text{(SPM)}}+\tau {{\sigma }^\text{(KA)}}
\end{equation}

\noindent where $\boldsymbol{p}$ is the 3D position where the ray hits any point on the surface element in the scene, $r$ is the distance between the object and the radar, $\theta $ is the angle between the ray and the target surface element. $\boldsymbol{\zeta}$ is the CSVBSDF parameters at any position, include parameters ${{\varepsilon }_{r}}$, $h$ and $l$, as shown in Table \ref{tab1}.

In this model, views each triangular facet is regarded as an independent rough microfacet, and numerous triangular facets form complex 3D targets or scenes. The scattering coefficients at any one location in each microrough surface are obtained by CSVBSDF. The CSVBSDF parameters can be learned by differentiable ray tracing, and the learned parameters can be used for high fidelity SAR image simulation.

\begin{figure}[!t]
\centering
\includegraphics[width=3.2in]{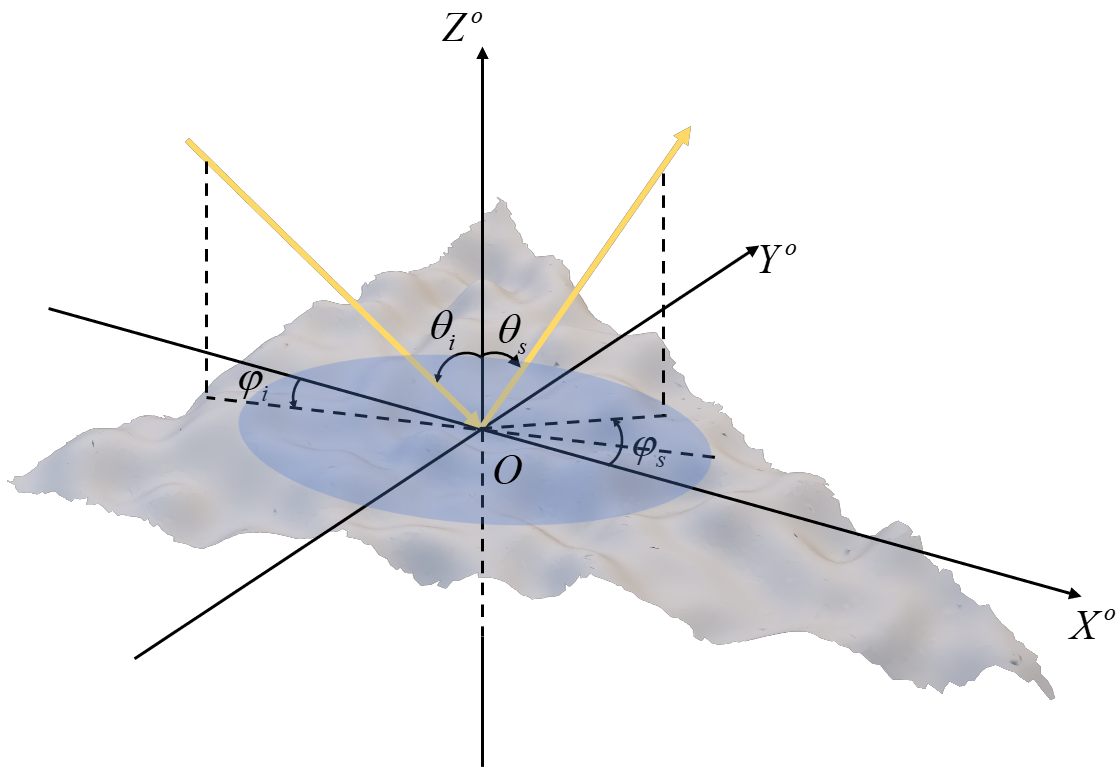}
\caption{Triangular rough microfacet scattering geometry diagram.}
\label{fig_2}
\end{figure}

Fig. \ref{fig_2} shows a schematic diagram of rough surface scattering of triangular surface elements. Table \ref{tab1}  lists the scattering parameters of a triangular rough surface element.

\begin{table}[!t]
\caption{SCATTERING MODEL VARIABLE NOTATIONS\label{tab:table1}}
\centering
\begin{tabular}{>{\rule[0ex]{0pt}{2ex}}c c}
\Xhline{1pt}  
Notation & Description  \\
\hline
${{\theta }_{i}}$ & Local incident zenith angle\\
\hline
${{\theta }_{s}}$ & Local scattering zenith angle\\
\hline
${{\varphi }_{i}}$ & Local incident azimuth angle\\
\hline
${{\varphi }_{s}}$ & Local scattering azimuth angle\\
\hline
$W(\cdot)$ & Rough surface power spectral density function\\
\hline
$l$ & Rough surface correlation length\\
\hline
$h$ & Rough surface root mean square height\\
\hline
${{\varepsilon }_{r}}$ & Relative dielectric constant\\
\Xhline{1pt}  
\end{tabular}
\label{tab1}
\end{table}

\subsubsection{Surface microwave specular BSDF}

In rough surface scattering calculations, KA is a commonly used high-frequency approximation method for processing specular scattering components. This method assumes that the radius of curvature of the rough surface is much larger than the wavelength of the incident electromagnetic wave, so that the rough surface can be locally regarded as a tangent plane. Therefore, the scattered field at any point on the rough surface can be approximately represented by the field on the tangent plane of the point. The KA backscattering coefficient of different polarizations can be expressed as:

\begin{equation}
\label{deqn_ex6}
\sigma _{hh}^\text{(KA)}=\sigma _{vv}^\text{(KA)}=\frac{|R(0){{|}^{2}}}{{{\cos }^{4}}{{\theta }_{i}}2{{h}^{2}}\left| {{C}^{\prime \prime }}(0) \right|}\exp \left( -\frac{{{\tan }^{2}}{{\theta }_{i}}}{2{{h}^{2}}\left| {{C}^{\prime \prime }}(0) \right|} \right)
\end{equation}.

\begin{equation}
\label{deqn_ex7}
\sigma _{hv}^\text{(KA)}=\sigma _{vh}^\text{(KA)}=0
\end{equation}

\noindent $R(0)$ is the Fresnel coefficient for zero angle of incidence:

\begin{equation}
\label{deqn_ex8}
R(0)=\frac{1-\sqrt{{{\varepsilon }_{r}}}}{1+\sqrt{{{\varepsilon }_{r}}}}.
\end{equation}

In Eq. (\ref{deqn_ex6}), ${{h}^{2}}\left| {{C}^{\prime \prime }}(0) \right|$ is the mean square surface slope, in the Gaussian related rough surface, ${{h}^{2}}\left| {{C}^{\prime \prime }}(0) \right|$ . The validity condition of KA can be summarized as:

\begin{equation}
\label{deqn_ex9}
{{k}_{1}}\ell >6,{{k}_{1}}\delta >\sqrt{10}/\left( \cos {{\theta }_{i}}-\cos {{\theta }_{s}} \right),{{R}_{c}}>\lambda.
\end{equation}

\begin{equation}
\label{deqn_ex10}
{{R}_{c}}={{\left. \delta \sqrt{\frac{2}{\pi }\frac{{{\partial }^{4}}C(\xi )}{\partial {{\xi }^{4}}}} \right|}_{\xi =0}}.
\end{equation}

\begin{equation}
\label{deqn_ex11}
{{k}_{1}}=(2\pi /\lambda )\sqrt{{{\varepsilon }_{1}}}.
\end{equation}

For Gaussian rough surfaces, it satisfy:

\begin{equation}
\label{deqn_ex12}
{{\ell }^{2}}>2.76\delta \lambda.
\end{equation}

The above constraints of Eq. (\ref{deqn_ex9}) and Eq. (\ref{deqn_ex12}) are used to constrain parameters during reverse differentiable training, so that the effective parameters of learning are kept within the valid range of the physical model.

\subsubsection{Surface microwave diffuse BSDF}

The perturbation method SPM is suitable for rough surfaces with small-scale undulations, and it expands the surface electromagnetic field according to the perturbation series of small parameters. SPM does not consider mirror reflection, and the scattering coefficient under the first-order perturbation approximation is:

\begin{equation}
\label{deqn_ex13}
\sigma _{pq}^\text{(SPM)}={{\alpha }^{2}}{{f}_{pq}}=8{{k}^{4}}{{\cos }^{2}}{{\theta }_{i}}{{\cos }^{2}}{{\theta }_{s}}W({{k}_{dx}},{{k}_{dy}}){{f}_{pq}}.
\end{equation}

\noindent where $\alpha $ is a function related to the rough surface, and ${{f}_{pq}}$ is the scattering amount related to the Muller matrix. $W$ is the PSD function of the rough surface, ${{k}_{dx}}$ and ${{k}_{dy}}$ are the wave numbers in the $x$ and $y$ directions respectively, satisfying the following equation: 

\begin{equation}
\label{deqn_ex14}
k_{dx}^{2}+k_{dy}^{2}={{k}^{2}}[{{\sin }^{2}}{{\theta }_{s}}+{{\sin }^{2}}{{\theta }_{i}}-2\sin {{\theta }_{s}}\sin {{\theta }_{i}}\cos ({{\varphi }_{s}}-{{\varphi }_{i}})].
\end{equation}

\noindent here, the description of ${{\theta }_{s}}$ ${{\theta }_{i}}$ ${{\varphi }_{s}}$ ${{\varphi }_{i}}$ are shown in Table \ref{tab1}.

\begin{figure}[t]
\centering
\includegraphics[width=3.5in]{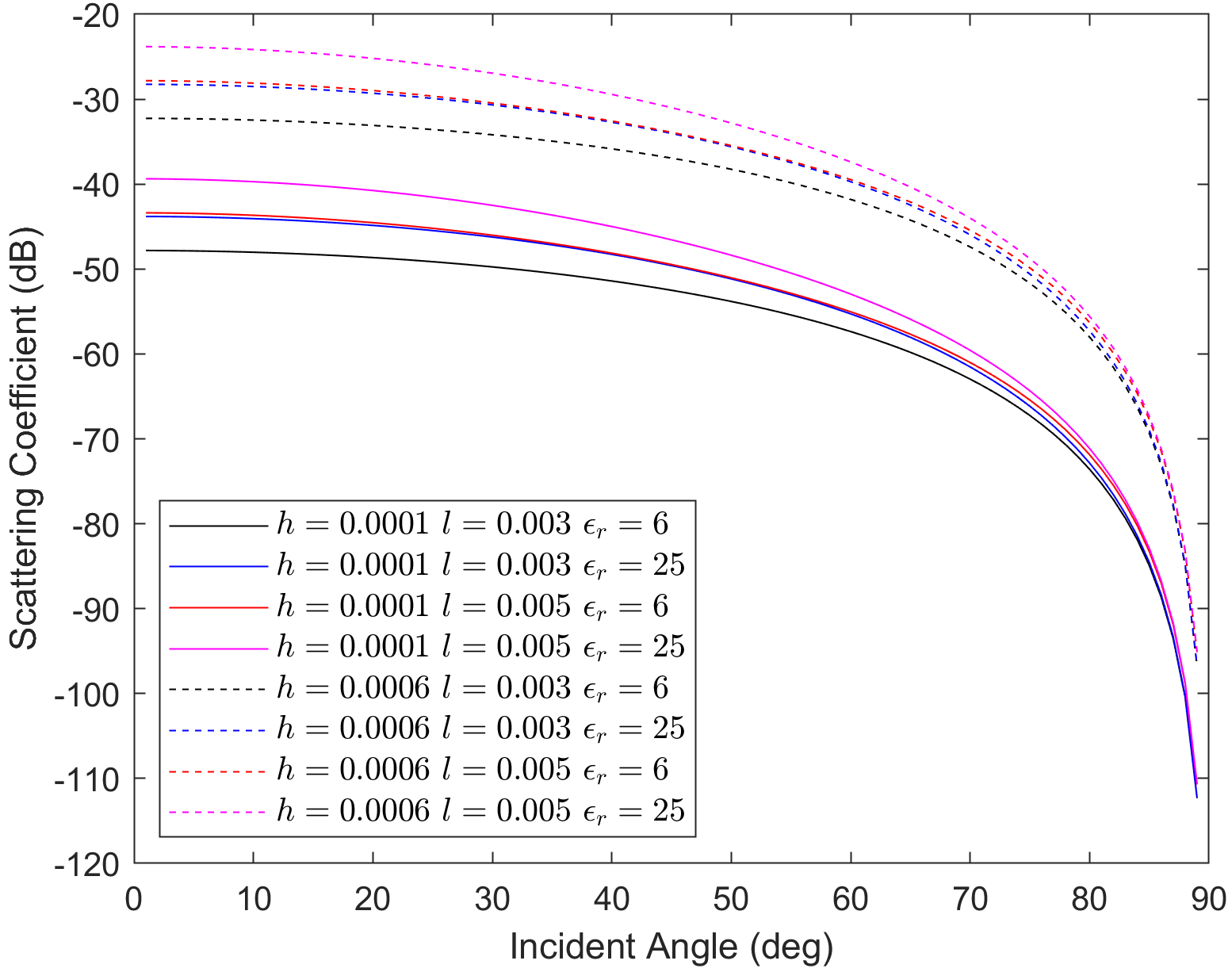}
\caption{Changes in scattering coefficient caused by different surface scattering parameters of SPM model.}
\label{fig_3}
\end{figure}

The Gaussian-correlated and exponential-correlated two-dimensional rough surface PSD are written as:

\begin{equation}
\label{deqn_ex15}
{{W}_{\text{Gauss}}}({{k}_{dx}},{{k}_{dy}})=\frac{{{h}^{2}}{{l}_{x}}{{l}_{y}}}{4\pi }\exp (-\frac{{{k}_{dx}}^{2}{{l}_{x}}^{2}+{{k}_{dy}}^{2}{{l}_{y}}^{2}}{4}).
\end{equation}

\begin{equation}
\label{deqn_ex16}
{{W}_{\text{Exp}}}({{k}_{dx}},{{k}_{dy}})=\frac{{{h}^{2}}{{l}_{x}}{{l}_{y}}}{{{\pi }^{2}}(1+{{k}_{dx}}^{2}{{l}_{x}}^{2})(1+{{k}_{dy}}^{2}{{l}_{y}}^{2})}.
\end{equation}

\noindent where ${{l}_{x}}$ and ${{l}_{y}}$ are the relative lengths of the rough surface in the  $x$ and $y$ directions, and ${h}$ as shown in Table \ref{tab1}.

${{f}_{pq}}$ in Eq. \ref{deqn_ex13} is exactly equal to the square of the Fresnel coefficient   under monostatic backscattered HH polarization:

\begin{equation}
\label{deqn_ex17}
{{f}_{hh}}={{\left( \frac{\cos \theta -\sqrt{{{\varepsilon }_{r}}-{{\sin }^{2}}\theta }}{\cos \theta +\sqrt{{{\varepsilon }_{r}}-{{\sin }^{2}}\theta }} \right)}^{2}}=R_{h}^{2}.
\end{equation}

\noindent where $\theta $ is the angle between the ray and the target surface element, ${{\varepsilon }_{r}}$ as shown in Table \ref{tab1}, and $f$ of VV polarization is:

\begin{equation}
\label{deqn_ex18}
{{f}_{vv}}={{\left( \frac{({{\varepsilon }_{r}}-1)({{\sin }^{2}}\theta -{{\varepsilon }_{r}}{{\cos }^{2}}\theta )}{{{({{\varepsilon }_{r}}\cos \theta +\sqrt{{{\varepsilon }_{r}}-{{\sin }^{2}}\theta })}^{2}}} \right)}^{2}}.
\end{equation}

It is worth noting that the scope of application of SPM is also limited. Only coherent scattering modeling is given here, and the effective application conditions are as follows:

\begin{equation}
\label{deqn_ex19}
kh\ll 1,{{k}^{3}}{{h}^{2}}l\ll 1,\sqrt{2}h/l<0.3.
\end{equation}

The constraints of the above Eq. \ref{deqn_ex19} are used to constrain parameters when performing inverse differentiable training. Fig. \ref{fig_3} shows the changes in scattering coefficient caused by different $h$, $l$, and ${{\varepsilon }_{r}}$ of the SPM model.

\begin{figure}[!t]
\centering
\includegraphics[width=3.5in]{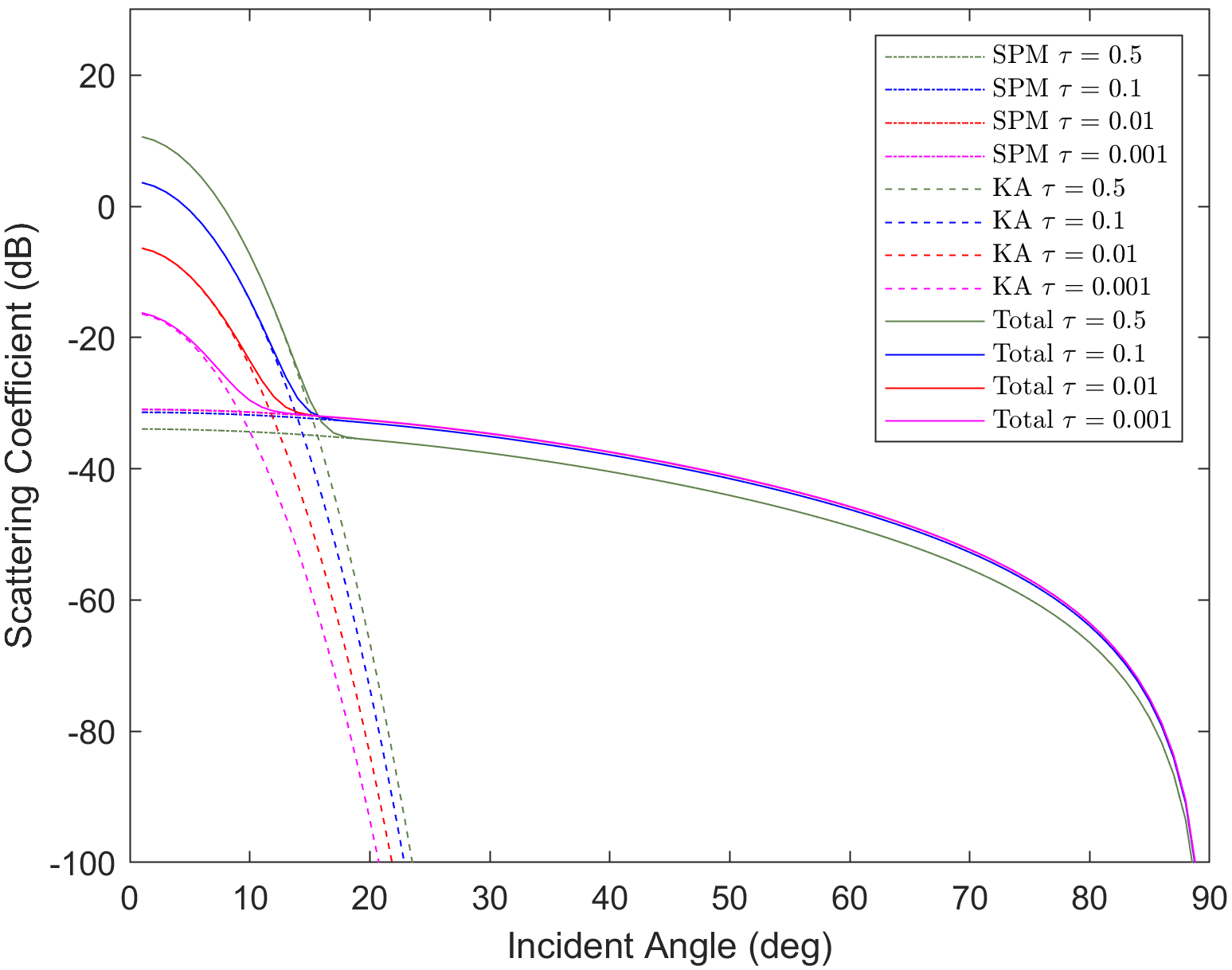}
\caption{Scattering coefficient fusion of double-scale model.}
\label{fig_4}
\end{figure}

\subsubsection{Surface Microwave BSDF}

Generally, KA is suitable for small angle $\left( <20{}^\circ  \right)$ incidence, corresponding to Specular BSDF item, and SPM is applicable for large angle $\left( \ge 20{}^\circ  \right)$ incidence, corresponding to Diffuse BSDF item. In real situations, the rough surface of a scene or object usually contains both rough parts and smooth parts. Therefore, integrating the two mechanisms in actual calculations will make the simulation results more realistic. 

In the actual forward and differentiable calculations, use the coefficient $\tau $ to adjust the ratio of ${{\sigma }^\text{(SPM)}}$ and ${{\sigma }^\text{(KA)}}$. Fig. \ref{fig_4} shows the  variation of scattering coefficient with $\tau $ in Eq. \ref{deqn_ex5}. 

\subsection{Ray-mesh intersection and surface parameters mapping}
\begin{figure}[!t]
\centering
\includegraphics[width=3.5in]{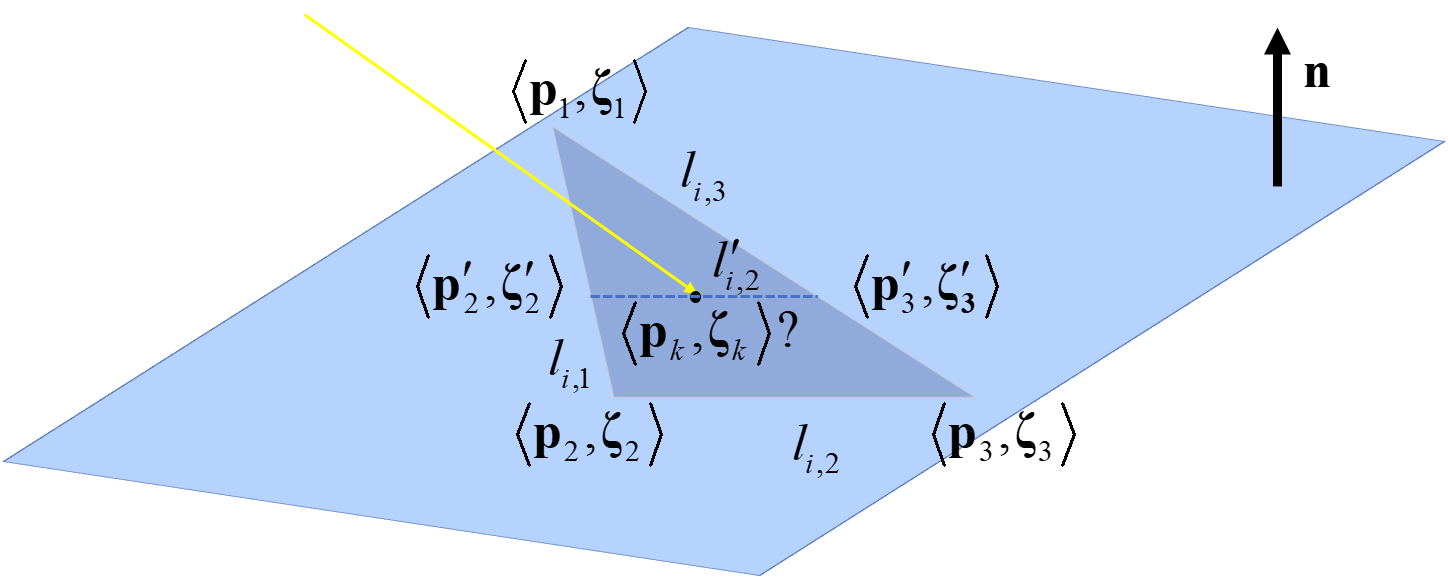}
\caption{Ray intersects triangle surface element.}
\label{fig_5}
\end{figure}

Three-dimensional meshes are used to represent target geometries, and the surface normal vectors are calculated from the triangular mesh vertices. As shown in Fig. \ref{fig_5}, for any ray defined as $\mathbf{p}(t)=\mathbf{o}+t\mathbf{d}$, $\mathbf{o}$ is the origin of the ray, $\mathbf{d}$ is the ray direction, $t\in [0,\infty )$ is a variable corresponding to the propagation distance. For any face element, there are three vertices ${{\mathbf{p}}_{1}}$, ${{\mathbf{p}}_{2}}$, ${{\mathbf{p}}_{3}}$, and the normal vector $\mathbf{n}$ within the element is defined as the unit vector in $\overrightarrow{{{\mathbf{p}}_{1}}{{\mathbf{p}}_{2}}}\times \overrightarrow{{{\mathbf{p}}_{1}}{{\mathbf{p}}_{3}}}$ that has a positive inner product with the ray $\mathbf{p}(t)$.

The intersection point of the ray and the surface element can be obtained through the ray equation and the plane equation. Any point ${{\mathbf{p}}_{k}}$ on the plane where the triangular surface element is located can be expressed by the barycentric coordinate equation:

\begin{equation}
\label{deqn_ex20}
{{\mathbf{p}}_{k}}=\left( 1-{{m}_{1}}-{{m}_{2}} \right){{\mathbf{p}}_{3}}+{{m}_{1}}{{\mathbf{p}}_{1}}+{{m}_{2}}{{\mathbf{p}}_{2}}.
\end{equation}

\noindent where $0 \leq {{m}_{1}} + {{m}_{2}} \leq 1$ are weight variables for vertices. A ray intersecting a plane can be expressed as:

\begin{equation}
\label{deqn_ex21}
\mathbf{o}+t\mathbf{d}=\left( 1-{{m}_{1}}-{{m}_{2}} \right){{\mathbf{p}}_{3}}+{{m}_{1}}{{\mathbf{p}}_{1}}+{{m}_{2}}{{\mathbf{p}}_{2}}.
 \end{equation}
 
\noindent where $t, {{m}_{1}}, {{m}_{2}}$ can be calculated by the following Eq.:

\begin{equation}
\label{deqn_ex22}
\begin{aligned}
&\begin{bmatrix}
  t \\
  m_{1} \\
  m_{2}
\end{bmatrix}
= \frac{1}{{\mathbf{f}_{1} \cdot \mathbf{h}_{1}}}
\begin{bmatrix}
   \mathbf{f}_{2} \cdot \mathbf{h}_{2} \\
  \mathbf{f}_{1} \cdot \mathbf{h} \\
  \mathbf{f}_{2} \cdot \mathbf{d}
\end{bmatrix}
\end{aligned}
\end{equation}

\noindent where $\mathbf{h}=\mathbf{o}-{{\mathbf{p}}_{3}}$, ${{\mathbf{h}}_{1}}={{\mathbf{p}}_{1}}-{{\mathbf{p}}_{3}}$, ${{\mathbf{h}}_{2}}={{\mathbf{p}}_{2}}-{{\mathbf{p}}_{3}}$, ${{\mathbf{f}}_{1}}=\mathbf{d}\times {{\mathbf{h}}_{2}}$, ${{\mathbf{f}}_{2}}=\mathbf{h}\times {{\mathbf{h}}_{1}}$. When $t>0,\text{ }{{m}_{1}}>0,\text{ }{{m}_{2}}>0$ is satisfied, the intersection point is within the triangle surface element, and the intersection point ${{\mathbf{p}}_{k}}$ can be obtained.

The surface scattering parameters of the CSVBSDF are represented using three large two-dimensional matrices, with coordinates following the UV coordinate system, which respectively represent the $h$, $l$, and ${{\varepsilon }_{r}}$ that affect roughness.

\begin{figure}[!t]
\centering
\includegraphics[width=3.5in]{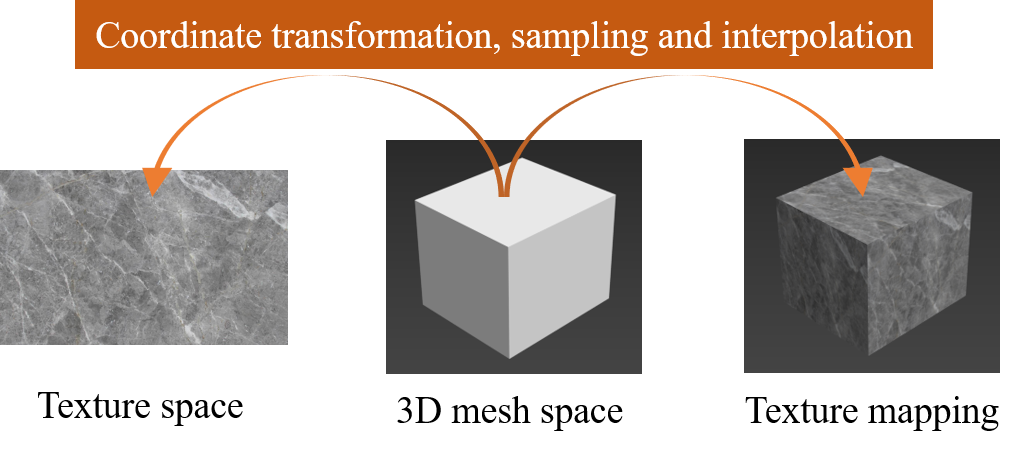}
\caption{Material and 3D mesh mapping.}
\label{fig_6}
\end{figure}

Visible light texture mapping establishes the mapping relationship between 3D mesh and texture space, as shown in Fig. \ref{fig_6}. Similarly, during SAR scattering simulation, we establish a two-way mapping relationship between 3D mesh and 2D BSDF surface scattering parameters. So that for a complex target, spatially varying BSDF parameters can be defined, such as roughness and dielectric constant. This mapping relationship can be realized through indexing 3D mesh and 2D scattering texture space, and the BSDF parameters within the surface element can be obtained through difference values. Each element in the surface scattering parameters matrix has a one-to-one mapping relationship with the fixed points of the triangular surface element in the three-dimensional geometry. The CSVBSDF parameters of any point in the surface element can be obtained by interpolating the parameters of the vertices.

As shown in Fig. \ref{fig_5}, the CSVBSDF parameters ${{\mathbf{\boldsymbol{\zeta} }}_{1}}$, ${{\mathbf{\boldsymbol{\zeta} }}_{2}}$, ${{\mathbf{\boldsymbol{\zeta} }}_{3}}$ of the vertex positions ${{\mathbf{p}}_{1}}$, ${{\mathbf{p}}_{2}}$, ${{\mathbf{p}}_{3}}$ can be obtained by retrieving the parameter list. There is the following proportional relationship:

\begin{equation}
\label{deqn_ex23}
\begin{aligned}
  & {{{\mathbf{{p}'}}}_{3}}-{{\mathbf{p}}_{k}}={{t}_{1}}\cdot ({{\mathbf{p}}_{3}}-{{\mathbf{p}}_{2}}),\text{       }  {{t}_{1}}\in \left[ 0,\infty  \right) \\ 
 & {{{\mathbf{{p}'}}}_{3}}-{{\mathbf{p}}_{1}}={{t}_{2}}\cdot ({{\mathbf{p}}_{3}}-{{\mathbf{p}}_{1}}),\text{       }  {{t}_{2}}\in \left[ 0,\infty  \right) \\ 
\end{aligned}
\end{equation}

Since the edge ${{{l}'}_{i,2}}$ is parallel to ${{l}_{i,2}}$, according to the similar triangle criterion:

\begin{equation}
\label{deqn_ex24}
\begin{aligned}
  & {{{\mathbf{{p}'}}}_{2}}-{{\mathbf{p}}_{1}}={{t}_{2}}\cdot ({{\mathbf{p}}_{2}}-{{\mathbf{p}}_{1}}) \\ 
 & {{{\mathbf{{p}'}}}_{3}}-{{{\mathbf{{p}'}}}_{2}}={{t}_{2}}\cdot ({{\mathbf{p}}_{3}}-{{\mathbf{p}}_{2}}) \\ 
\end{aligned}
\end{equation}

According to the triangle vector relationship:

\begin{equation}
\label{deqn_ex25}
{{\mathbf{{p}'}}_{3}}-{{\mathbf{p}}_{1}}=\left( {{\mathbf{p}}_{k}}-{{\mathbf{p}}_{1}} \right)+\left( {{{\mathbf{{p}'}}}_{3}}-{{\mathbf{p}}_{k}} \right).
\end{equation}

Substituting Eq. (\ref{deqn_ex23}) and Eq. (\ref{deqn_ex24}) into Eq. (\ref{deqn_ex25}):

\begin{equation}
\label{deqn_ex26}
{{t}_{2}}\cdot ({{\mathbf{p}}_{3}}-{{\mathbf{p}}_{1}})={{\mathbf{p}}_{k}}-{{\mathbf{p}}_{1}}+{{t}_{1}}\cdot ({{\mathbf{p}}_{3}}-{{\mathbf{p}}_{2}}).
\end{equation}

\noindent ${{t}_{1}}$ and ${{t}_{2}}$ can be obtained through above equation, and the material parameter ${{\mathbf{{\boldsymbol{\zeta} }'}}_{1}}$, ${{\mathbf{{\boldsymbol{\zeta} }'}}_{2}}$ of point ${{\mathbf{{p}'}}_{2}}$, ${{\mathbf{{p}'}}_{3}}$ can be obtained through linear interpolation:

\begin{equation}
\label{deqn_ex27}
\begin{aligned}
  & {{{\mathbf{{\boldsymbol{\zeta} }'}}}_{2}}=\left( 1-{{t}_{2}} \right)\cdot {{\mathbf{\boldsymbol{\zeta} }}_{1}}+{{t}_{2}}\cdot {{\mathbf{\boldsymbol{\zeta} }}_{2}} \\ 
 & {{{\mathbf{{\boldsymbol{\zeta} }'}}}_{3}}=\left( 1-{{t}_{2}} \right)\cdot {{\mathbf{\boldsymbol{\zeta} }}_{1}}+{{t}_{2}}\cdot {{\mathbf{\boldsymbol{\zeta} }}_{3}} \\ 
\end{aligned}
\end{equation}

Assume that the proportional relationship between ${{\mathbf{p}}_{k}}$ and ${{\mathbf{{p}'}}_{2}}$, ${{\mathbf{{p}'}}_{3}}$ is:

\begin{equation}
\label{deqn_ex28}
{{\mathbf{{p}'}}_{3}}-{{\mathbf{p}}_{k}}=t\cdot ({{\mathbf{{p}'}}_{3}}-{{\mathbf{{p}'}}_{2}}),\text{          }t\in \left[ 0,\infty  \right).
\end{equation}

Substituting Eq. (\ref{deqn_ex23}) and Eq. (\ref{deqn_ex24}) into Eq. (\ref{deqn_ex28}), $t=\left| \frac{{{t}_{1}}}{{{t}_{2}}} \right|$ can be calculated. The CSVBSDF surface scattering parameters ${{\mathbf{\boldsymbol{\zeta} }}_{k}}$ of the hit point can be further calculated:

\begin{equation}
\label{deqn_ex29}
{{\mathbf{\boldsymbol{\zeta} }}_{k}}=\left( 1-t \right)\cdot {{\mathbf{{\boldsymbol{\zeta} }'}}_{3}}+t\cdot {{\mathbf{{\boldsymbol{\zeta} }'}}_{2}}.
\end{equation}

\begin{figure*}[htbp]
\centering
\includegraphics[width=7in]{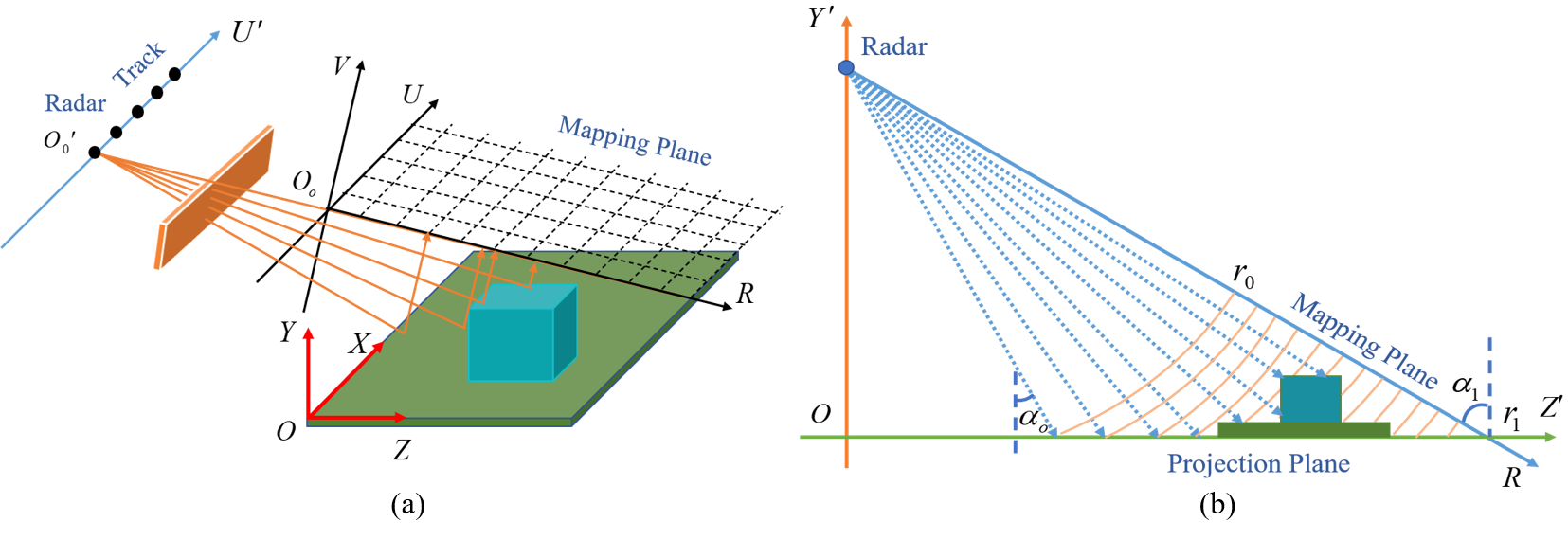}
\caption{SAR mapping and projected and definition of related coordinate systems. (a) is 3D illustration, (b) is 2D illustration at a certain azimuth bin.}
\label{fig_7}
\end{figure*}

\subsection{Mapping and projection for fast SAR imaging}
After calculating the scattering intensity of each ray, mapping and projection are performed based on the SAR imaging principle, and fast imaging is performed on the mapping plane, as shown in Fig. \ref{fig_7}. 

The radar distinguishes targets according to different distances. The scattering of targets with the same distance is mapped to the same pixel. After the radar emits rays, the scattering amount of each position in the scene is based on the distance from the radar. They are respectively mapped to the mapping plane, that is, the plane on the axis in the figure. The total scattering coefficient of a row can be obtained by summing up the scattering of target points at equal distances, where the mapping plane is divided equally according to the distance resolution. According to the size of the scene and the position of the radar, the sampling range can be determined, the slant distance $r\in [{{r}_{0}},{{r}_{1}}]$, and the angle of incidence $\alpha \in [{{\alpha }_{0}},{{\alpha }_{1}}]$.

As shown in Fig. \ref{fig_7}(a), $O-XYZ$ is the world coordinate system. The $O{{'}_{0}}U'$ axis is the tangent direction of the radar motion trajectory. During the motion, rays with electromagnetic wave information are continuously emitted. The radar emits an electromagnetic wave ray array at the sampling position. When the distance between the radar and the target is far, the emitted electromagnetic waves can be regarded as plane waves. Fig. \ref{fig_7}(b) is a schematic diagram of SAR mapping projection imaging at a certain azimuth bin.
	
A point ${{O}_{0}}$ on the top ray of the sector ray cluster emitted from the initial position of the radar is selected as the origin of the mapping coordinate system. The vector ${{O}_{0}}U$ is consistent with the radar motion trajectory direction $O{{'}_{0}}U'$, and the vector ${{O}_{0}}R$ is consistent with the top ray direction of the fan-shaped ray cluster emitted from the initial position of the radar. ${{O}_{0}}V$ is obtained by the cross product of ${{O}_{0}}U$ and ${{O}_{0}}R$. From this, the mapping coordinate system ${{O}_{0}}-UVR$ can be constructed. After the incident wave hits the 3D target, the scattering information of each ray is calculated and the hit point data is recorded, and the coordinate system of the hit point data is converted.
\textbf{}

Affine transformation is used to transform the hit point and scattering information in the world coordinate system to the mapping coordinate system. The transformation equation between the world coordinate system and the mapping coordinate system is as follows:

\begin{equation}
\label{deqn_ex30}
{{\mathbf{p}}_{m}}={{\mathbf{R}}_{m}}{{\mathbf{p}}_{w}}+\mathbf{T}.
\end{equation}
 
\noindent here ${{\mathbf{p}}_{w}}$ is the coordinate of the hit point in the world coordinate system, and $\mathbf{T}$ is the translation matrix between the mapping coordinate system ${{O}_{0}}-UVR$ and the world coordinate system $O-XYZ$. ${{\mathbf{p}}_{m}}$ is the hit point coordinates in the mapping coordinate system, ${{\mathbf{R}}_{m}}$ is the rotation matrix from the world coordinate system $O-XYZ$ to the mapping coordinate system ${{O}_{0}}-UVR$.

\begin{equation}
\label{deqn_ex31}
{{\mathbf{R}}_{m}}=
\begin{bmatrix}
  -\cos \beta & -\cos \gamma \sin \beta & -\sin \gamma \sin \beta \\
  0 & \sin \gamma & -\cos \gamma \\
  \sin \beta & -\cos \gamma \cos \beta & -\sin \gamma \cos \beta
\end{bmatrix}.
\end{equation}

\begin{table}[htbp]
\caption{MAPPING AND PROJECTION IMAGING VARIABLE NOTATIONS\label{tab:table2}}
\centering
\begin{tabular}{>{\centering\arraybackslash} >{\rule[0ex]{0pt}{2ex}}m{2cm} 
>{\centering\arraybackslash}m{5cm}}
\Xhline{1pt}  
Notation & Description  \\
\hline
${{\mathbf{p}}_{w}}$ & The ray hit points coordinates in the world coordinate system\\
\hline
${{\mathbf{p}}_{m}}$ & Hit point coordinates in the mapping coordinate system\\
\hline
${{R}_{u}}$ & Azimuth resolution unit\\
\hline
${{R}_{r}}$ & Range resolution unit\\
\hline
${{\mathbf{H}}_{r}}$ & The coordinates of ${{\mathbf{p}}_{m}}$ of a ray cluster along the ${{O}_{0}}R$ direction\\
\hline
$\mathbf{H}_{r}^{'}$ & The vector after ${{\mathbf{H}}_{r}}$ is sorted\\
\hline
$\mathbf{H}_{f}^{'}$ & $\mathbf{H}_{r}^{'}$ rounded by resolution ${{R}_{r}}$\\
\hline
$\mathbf{k}$ & ${{\mathbf{H}}_{r}}$ sorted index\\
\hline
${{\mathsf{\mathcal{I}}}_{r}}$ & The scattering intensity of the ray cluster at all hit points from incident angle ${{\alpha }_{0}}$ to ${{\alpha }_{1}}$\\
\hline
$\mathcal{I}$ & Simulated SAR image\\
\Xhline{1pt}  
\end{tabular}
\label{tab2}
\end{table}

$\gamma $ is the relative pitch angle of ${{O}_{0}}-UVR$ and $O-XYZ$, $\beta $ is the azimuth angle. Then the coordinates of each hit point under the mapping plane are calculated, and finally the radiation amount of each hit point in the scene is accumulated to the mapping plane at the same distance.

The backscattering ${{k}_{dx}}$ and ${{k}_{dy}}$ of the diffuse BSDF term satisfy:

\begin{equation}
\label{deqn_ex33}
k_{dx}^{2}+k_{dy}^{2}=4{{\sin }^{2}}\theta {{k}^{2}}.
\end{equation}

Integrating the radar along the equal slope line can obtain the sum of all scattering energy on the equal slope line, and the scattering intensity ${{\mathcal{I}}_{i,j}}$ of pixel  $(i,j)$ with the number of ray samples $\mathcal{M}$ can be expressed as:

\begin{equation}
\label{deqn_ex34}
{{\mathsf{\mathcal{I}}}_{i,j}}\text{=}\int_{{{u}_{n}}}^{{{u}_{n+1}}}{\operatorname{d}u}\int_{{{r}_{n}}}^{{{r}_{n+1}}}{\operatorname{d}r}\int_{{{\theta }_{0}}}^{{{\theta }_{1}}}{\frac{1}{\mathcal{M}}\sum\limits_{m=0}^{\mathcal{M}}{{{\mathcal{S}}_{m}}(p,r,\theta ,\mathsf{\mathcal{Z}})}\operatorname{d}\theta }.
\end{equation}

\noindent where $i\in [{{u}_{n}},{{u}_{n+1}})$, $j\in [{{r}_{v}},{{r}_{v+1}})$, ${{u}_{n}}=n{{R}_{u}}$, ${{r}_{v}}=v{{R}_{r}}+{{r}_{0}}$. ${{R}_{u}}$ and ${{R}_{r}}$ are the resolution units in the azimuth and range directions. $\theta \in [{{\theta }_{0}},{{\theta }_{1}})$ is the incident angle of the ray in each pixel unit. Therefore, the scattering intensity distribution SAR image under the mapping plane coordinate system ${{O}^{'}}-UVR$ can be obtained by traversing each ray. Although the scattering intensity of each ray can be quickly obtained through parallel computing, it is very time-consuming to traverse each ray to map it in actual projection mapping imaging. We designed a fast algorithm for mapping and projection calculate and ensure it is differentiable.

The radar emits a ray grid at a certain moment ${{u}_{n}}$, as shown in Fig. \ref{fig_7}(a). Note that the 3D position of all hit points of the ray grid from the incident angle ${{\alpha }_{0}}$ to ${{\alpha }_{1}}$ in the mapping coordinate system ${{O}_{0}}-UVR$ is $\mathbf{H}$, and the coordinates along the ${{O}_{0}}R$ direction are marked as ${{\mathbf{H}}_{r}}$. Sorting ${{\mathbf{H}}_{r}}$ is denoted as $\mathbf{H}_{r}^{'}$, and the sorted index is denoted as $\mathbf{k}$. Round $\mathbf{H}_{r}^{'}$ according to the resolution of ${{R}_{r}}$ and denote it as $\mathbf{H}_{f}^{'}$. Table \ref{tab2} summarizes the relevant variables for projection mapping imaging.

\begin{equation}
\label{deqn_ex35}
\mathbf{H}_{f}^{'}=\left[ \frac{\max (\mathbf{H})-\mathbf{H}_{r}^{'}}{{{R}_{r}}} \right].
\end{equation}

The scattering intensity of all hit points of the ray grid from the incident angle ${{\alpha }_{0}}$ to ${{\alpha }_{1}}$ is recorded as ${{\mathsf{\mathcal{I}}}_{r}}$, and ${{\mathsf{\mathcal{I}}}_{r}}$ is reorganized according to the index $\mathbf{k}$, which is recorded as $\mathsf{\mathcal{I}}_{r}^{'}$. Finally the $\mathsf{\mathcal{I}}_{r}^{'}$ elements corresponding to the same value in $\mathbf{H}_{f}^{'}$ are accumulated to obtain the SAR one-dimensional range image at time ${{u}_{n}}$. Repeating this operation for any time, can obtain the two-dimensional SAR image ${{\mathsf{\mathcal{I}}}_\text{sar}}$.

\section{Differentiable Ray Tracing}

The forward model in the previous section can be seen as a complex function \(\mathcal{I}\) affected by input parameters. These input parameters include radar observation parameters, target geometry parameters and CSVBSDF parameters \(\boldsymbol{\zeta} \), etc. This section builds an inverse model based on differentiable ray tracing. According to the difference between the SAR images simulated by the forward model and the reference images, the inverse learning of parameter \(\boldsymbol{\zeta} \) is performed through DRT.

\subsection{Gradient descent parameter learning}

In this part, the CSVBSDF parameter $\boldsymbol{\zeta}$ is optimized by minimizing the loss function $\mathcal{L}$ through gradient backpropagation:

\begin{equation}
\label{deqn_ex36}
{{\boldsymbol{\zeta} }^{*}}=\arg \min \mathcal{L}\left( \left( \mathcal{I}\left( \boldsymbol{\zeta}  \right),\overset{\sim}{\mathop{\mathcal{I}}}\, \right);\boldsymbol{\zeta}  \right).
\end{equation}

\noindent where $\mathcal{I}\left( \boldsymbol{\zeta}  \right)$ is a set of simulation images generated by the object geometry and CSVBSDF parameter $\boldsymbol{\zeta}$, $\overset{\sim}{\mathop{\mathsf{\mathcal{I}}}}$ is a set of reference measured SAR images, and $\mathcal{L}$ is the loss function of the simulation image $\mathcal{I}$ and the measured image $\overset{\sim}{\mathop{\mathcal{I}}}$, see Eq. (\ref{deqn_ex37}), Eq. (\ref{deqn_ex38}) and Eq. (\ref{deqn_ex39}) for details.

\subsubsection{Loss function}

The CSVBSDF parameter at any position consists of $\boldsymbol{\zeta} =(h,l,{{\varepsilon }_{r}})$, which is represented by three large two-dimensional matrices. Gradient backpropagation can be performed through the loss function of the simulated image ${{\mathsf{\mathcal{I}}}_\text{sar}}$ and the measured image $\overset{\sim}{\mathop{\mathsf{\mathcal{I}}}}$, and $\overset{\sim}{\mathop{\mathsf{\mathcal{I}}}}\,=\left( {{\overset{\sim }{\mathop{I}}\,}_{1}},{{\overset{\sim }{\mathop{I}}\,}_{2}},{{\overset{\sim }{\mathop{I}}\,}_{3}},\cdot \cdot \cdot  \right)$ is a set of observation images of the target. Therefore, the loss function is:

\begin{equation}
\label{deqn_ex37}
\mathcal{L}\left( \left( \mathcal{I}\left( \boldsymbol{\zeta}  \right),\overset{\sim}{\mathop{\mathcal{I}}}\, \right);\boldsymbol{\zeta}  \right)={{\mathcal{L}}_\text{sim}}(\mathsf{\mathcal{I}}\left( \boldsymbol{\zeta}  \right);\overset{\sim}{\mathop{\mathsf{\mathcal{I}}}}\,)+{{\mathcal{L}}_\text{mat}}\left( \boldsymbol{\zeta}  \right).
\end{equation}

\noindent here ${{\mathcal{L}}_\text{sim}}$ represents the loss between the image simulated by the simulator and the actual observed SAR image, and ${{\mathcal{L}}_\text{mat}}\left( \boldsymbol{\zeta}  \right)$ represents the regularization constraint of the CSVBSDF parameters, which is used to improve the robustness of the optimization. ${{\mathcal{L}}_\text{sim}}$ is expressed as follows:

\begin{equation}
\label{deqn_ex38}
\begin{aligned}
{{\mathsf{\mathcal{L}}}_\text{sim}}\left( \mathsf{\mathcal{I}}\left( \boldsymbol{\zeta}  \right);\overset{\sim}{\mathop{\mathsf{\mathcal{I}}}}\, \right)
&=\frac{{{\lambda }_\text{sim}}}{U\times M\times N} \\
&\times \sum\limits_{k=1}^{U}{\sum\limits_{i=1}^{M}{\sum\limits_{j=1}^{N}{{{({{\mathsf{\mathcal{I}}}_\text{sar}}(i,j)-\overset{\sim }{\mathop{\mathsf{\mathcal{I}}}}\,(i,j))}^{2}}}}}.
\end{aligned}
\end{equation}

\noindent where ${\mathit{\lambda }_\text{sim}}$ is a specified weight, ${{\mathsf{\mathcal{I}}}_\text{sar}}$ is the forward simulated SAR image, $U$ is the number of a group of images participating in differentiable training, $M,N$ is the width and height of a SAR image respectively.
In order to standardize the optimization of CSVBSDF surface scattering parameters, ${{\mathcal{L}}_\text{mat}}\left(\boldsymbol{\zeta} \right)$ is introduced to make the spatial optimization smoother. This prevents steep transitions and outliers in the optimization parameters, which can be easily obtained through automatic differentiation, expressed as:

\begin{equation}
\label{deqn_ex39}
\begin{aligned}
{{\mathcal{L}}_{\text{mat}}}\left( \boldsymbol{\zeta}  \right) 
&= {{\lambda }_{\text{mat}}}\sum\nolimits_{h,w,i}\left( \left| \boldsymbol{\zeta} [h+1,w,i]-\boldsymbol{\zeta} [h,w,i] \right| \right.  \\
&\quad + \left. \left| \boldsymbol{\zeta} [h,w+1,i]-\boldsymbol{\zeta} [h,w,i] \right| \right).
\end{aligned}
\end{equation}

\noindent where ${{\lambda }_\text{mat}}$ is the specified weight, $\boldsymbol{\zeta} [h,w,i]$ is the $(h,w)$ element parameter of the $i$-th type, $i=0$ represents $h$, $i=1$ represents $l$, and $i=2$ represents ${{\varepsilon }_{r}}$.

\begin{figure}[t]
\centering
\includegraphics[width=3.5in]{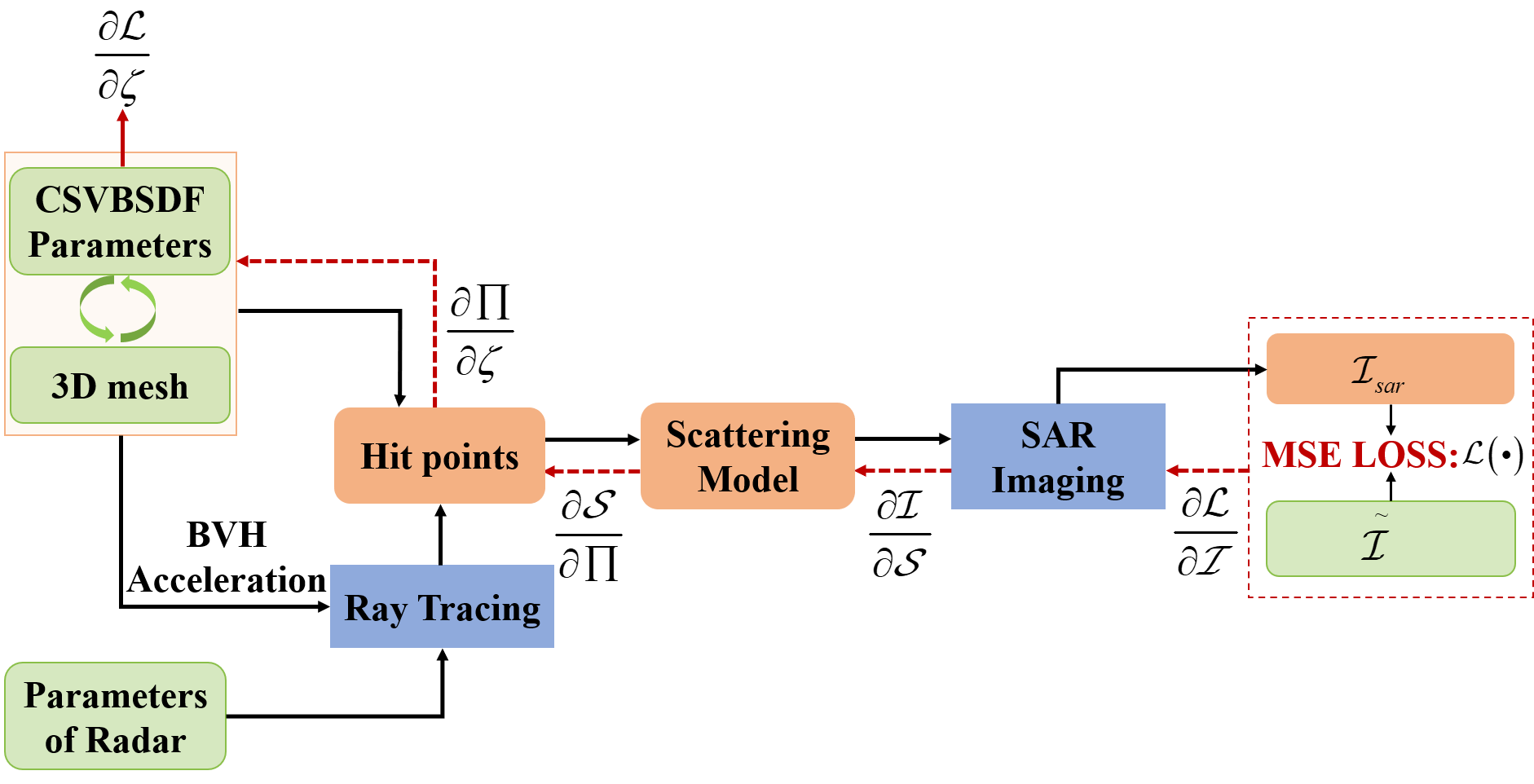}
\caption{Gradient backpropagation learning process. ${\partial \mathsf{\mathcal{L}}}/{\partial \mathcal{I}}$ represents the gradient of the loss function to the simulated SAR image ${{\mathsf{\mathcal{I}}}_\text{sar}}$, ${\partial \mathsf{\mathcal{I}}}/{\partial \mathcal{S}}$ represents the derivative of the simulated image to the scattering intensity, ${\partial \mathcal{S}}/{\partial \prod }$ represents the derivative of the scattering intensity to the ray hit point, ${\partial \prod }/{\partial \boldsymbol{\zeta} }$ represents the derivative of the ray hit point to the CSVBSDF parameters.}
\label{fig_8}
\end{figure}

\subsubsection{Gradient backpropagation}

After determining the correspondence between the simulated SAR image and the parameters to be optimized. According to the chain derivation rule, the gradient is gradually calculated, the gradient of the CSVBSDF parameters can be calculated as follows:

\begin{equation}
\label{deqn_ex40}
\frac{\text{d}\mathsf{\mathcal{L}}}{\text{d}\boldsymbol{\zeta} }=\frac{\partial \mathcal{L}}{\partial \mathcal{I}}\frac{\text{d}\mathcal{I}}{\text{d}\boldsymbol{\zeta} }+\frac{\partial \mathcal{L}}{\partial \boldsymbol{\zeta} }.
\end{equation}

\noindent where ${\partial \mathsf{\mathcal{L}}}/{\partial \mathcal{I}}$ and ${\partial \mathsf{\mathcal{L}}}/{\partial \boldsymbol{\zeta} }$ can be computed by automatic differentiation, and estimating the gradient of ${\text{d}\mathcal{I}}/{\text{d}\boldsymbol{\zeta} }$ requires performing differentiable ray tracing simulations. Each time a forward simulation is completed, the gradient is calculated and backward propagated to learning the CSVBSDF parameters.

During the forward simulation process, the intermediate values are recorded for the gradient calculation of back propagation. Fig. \ref{fig_8} shows the gradient back propagation optimization process, and the chain derivation is performed through the loss function of Eq. (\ref{deqn_ex37}). After the end of a differentiable simulation process, ${\partial \mathsf{\mathcal{L}}}/{\partial \boldsymbol{\zeta} }$ can be obtains to optimize the parameter $\boldsymbol{\zeta}$.

\subsection{Implementation}
The verification experiment was conducted based on NVIDIA Quadro RTX 8000 graphics card, and OptiX \cite{parker2010optix} was used to perform ray-triangle intersection query. Intersection acceleration calculations are performed using the Boundary Volume Hierarchy (BVH). To obtain a more accurate pixel for scatter intensity, upsampling is performed around each ray. A more accurate scattering quantity can be obtained by hemispheric integration achieved by Monte Carlo sampling \cite{li2018differentiable}. When calculating backscatter, uniform Monte Carlo sampling is performed according to Eq. (\ref{deqn_ex34}). Bi-station scattering can improve accuracy through importance sampling. Note that OptiX is not aware of differentiable scene parameters, so the generated intersection information must be converted into an intersection with differentiable position, normal, tangent, and CSVBSDF parameter coordinates.

\begin{algorithm}[t]
\caption{ Learning Surface Scattering Parameters}\label{alg:alg1}
\begin{algorithmic}
\STATE 
\STATE {\textbf{Input:} \text{ $\overset{\sim}{\mathop{\mathsf{\mathcal{I}}}}\,=\left( {{\overset{\sim }{\mathop{I}}\,}_{1}},{{\overset{\sim }{\mathop{I}}\,}_{2}},{{\overset{\sim }{\mathop{I}}\,}_{3}},\cdot \cdot \cdot  \right)$, $\delta $ and object 3D mesh}}
\STATE {\text{Initialization: $\boldsymbol{\zeta} \leftarrow (h,l,{{\varepsilon }_{r}})$, $\partial \boldsymbol{\zeta} $}}
\STATE {\text{\% Generate large-scale ray arrays in parallel}}
\STATE {\textbf{for} \text{${{\mathbf{r}}_{i}}\leftarrow {{\mathbf{r}}_{0}}$ to ${{\mathbf{r}}_{1}}$ and ${{\alpha }_{j}}\leftarrow {{\alpha }_{0}}$ to ${{\alpha }_{1}}$ \textbf{do}}}
\STATE \hspace{0.5cm} \text{Calculate $\prod$ of ray-mesh intersection by BVH.}
\STATE \hspace{0.5cm} \text{Calculate $\boldsymbol{\zeta} \leftarrow \prod$}\
\STATE \hspace{0.5cm} \text{Calculate $\mathcal{S}(p,r,\theta , \boldsymbol{\zeta} )$ based on Eq. (\ref{deqn_ex5})}
\STATE \hspace{0.5cm} \text{Calculate ${{\mathsf{\mathcal{I}}}_{i,j}}$ based on Eq. (\ref{deqn_ex34})}
\STATE \hspace{0.5cm} \text{${{\mathsf{\mathcal{I}}}_\text{sar}}\leftarrow {{\mathsf{\mathcal{I}}}_{i,j}}$}
\STATE \textbf{end for}
\STATE {\textbf{for} \text{${{\mathbf{r}}_{i}}\leftarrow {{\mathbf{r}}_{0}}$ to ${{\mathbf{r}}_{1}}$ and ${{\alpha }_{j}}\leftarrow {{\alpha }_{0}}$ to ${{\alpha }_{1}}$ \textbf{do}}}
\STATE \hspace{0.5cm} \text{Calculate $\mathcal{L}\left( \left( \mathsf{\mathcal{I}}\left( \boldsymbol{\zeta}  \right),\boldsymbol{\zeta}  \right);\overset{\sim}{\mathop{\mathsf{\mathcal{I}}}}\, \right)$ based on Eq. (\ref{deqn_ex36})}
\STATE \hspace{0.5cm} \text{Calculate $\partial \mathsf{\mathcal{I}}$, $\partial \mathcal{S}$ and $\partial \prod$.}
\STATE \hspace{0.5cm} \text{Calculate $\partial \boldsymbol{\zeta}$ based on Eq. (\ref{deqn_ex40})}
\STATE \textbf{end for}
\STATE \textbf{Update $\boldsymbol{\zeta} \leftarrow \boldsymbol{\zeta} -\partial \boldsymbol{\zeta}$}
\STATE \textbf{Output: $\boldsymbol{\zeta}$}
\end{algorithmic}
\label{alg1}
\end{algorithm}

According to automatic differentiation, it is easy to combine various differentiable scene parameters, such as bin vertex positions, transformation matrices of radar and scene, scattering rate, and CSVBSDF parameters such as permittivity, roughness, etc.

The initial position of the radar is ${{\mathbf{r}}_{0}}$, the end position is ${{\mathbf{r}}_{1}}$, and any position in the middle is ${{\mathbf{r}}_{i}}$. Rays from incident angles ${{\alpha }_{0}}$ to ${{\alpha }_{1}}$ are emitted from any position, and any incident angle in the middle is ${{\alpha }_{j}}$. Note the radar observation parameter is $\delta $.

DRT takes as input an initial configuration of target surface scattering parameters. Using Adam's algorithm to minimize the loss of Eq. (\ref{deqn_ex37}). The optimized results can be used as input for the next simulation.

\section{Experiments}
This section conducts experimental verification of DRT parameter learning. Two experiments are designed. The first subsection is BSDF parameter learning based on simulated SAR images. Concretely, the reference images are simulated SAR images, and the learned parameters are unified and non-spatially varying. The second subsection is the CSVBSDF parameter learning based on the measured SAR images. The reference images are the measured SAR image. The parameters learned are spatially variable, and the plane scattering parameters of each surface element can be learned.

\subsection{Learning BSDF parameters based on simulated SAR images}
In this subsection, experiments are designed based on simulated SAR images to achieve differentiable inversion learning of surface scattering parameters. Perform ray tracing and mapping projection forward simulation on targets with known geometry, and select material parameters for different media based on experience.

\begin{figure}[t]
\centering
\includegraphics[width=3.5in]{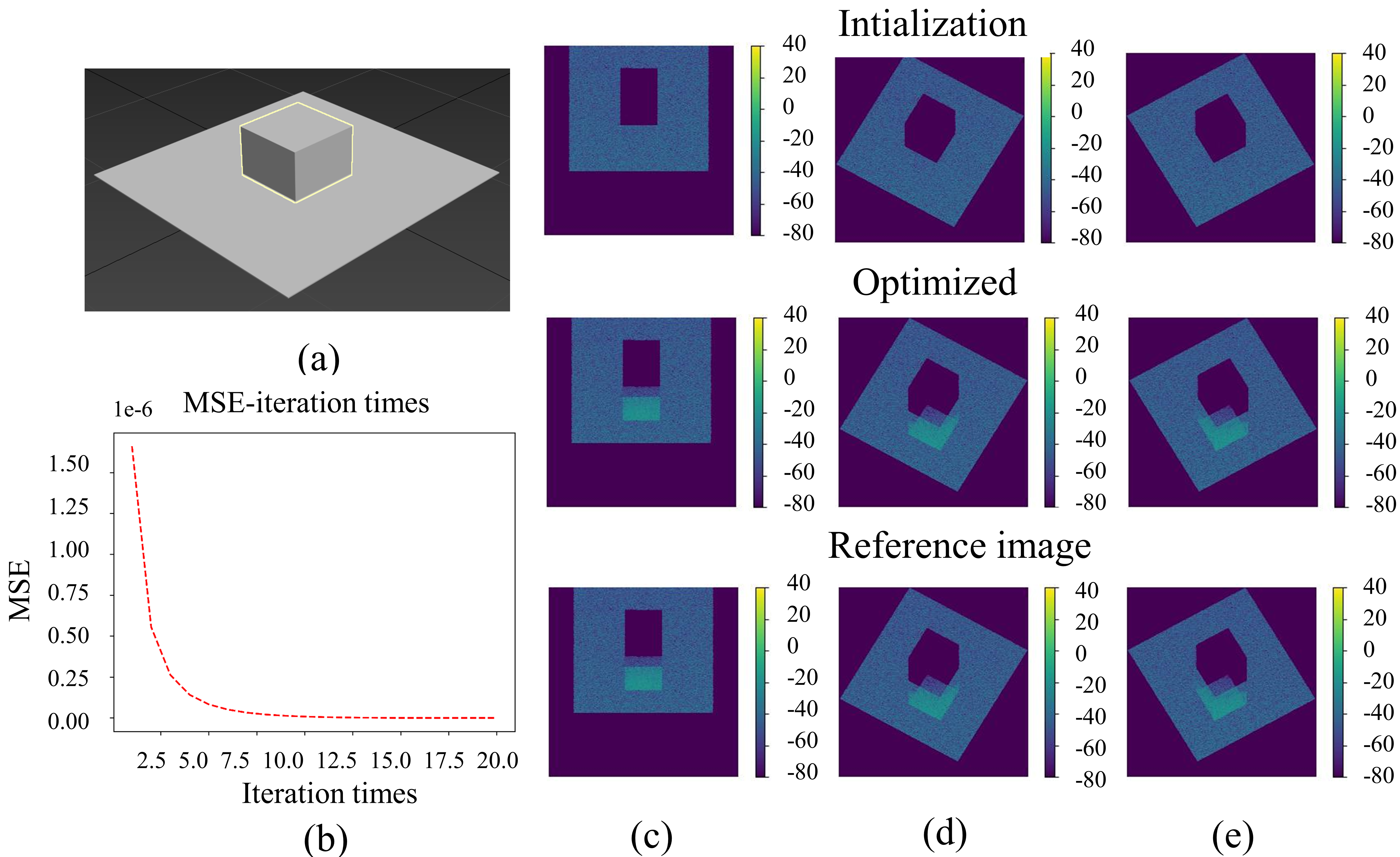}
\caption{SAR image simulation and differentiable optimization results in simple scenarios. (a) is the designed simple block 3D scene; (b) is the curve of the loss value of DRT differentiable training with iteration times; (c) (d) (e) is the initial, optimized and reference SAR images from three different azimuth.}
\label{fig_9}
\end{figure}

\begin{table}[t]
\caption{Cube material parameter optimization and comparison with Ground Truth\label{tab:table3}}
\centering
\begin{tabular}{>{\rule[0ex]{0pt}{2ex}}c c c c}
\Xhline{1pt}  
Parameters & Initialization & Optimized & Ground Truth \\
\hline
${{\varepsilon }_{r}}$ & 25 & 73.95 & 75\\
\hline
$h$(m) & 0.005 & 0.00209 & 0.002\\
\hline
$l$(m) & 0.01 & 0.00104 & 0.001\\
\Xhline{1pt}  
\end{tabular}
\label{tab3}
\end{table}

As shown in Fig. \ref{fig_9}(a), we set up a simple experimental scene, which consists of a plane and a cube. The setting parameters of the scene are as follows: the center frequency is 9.6GHz, and the incident angle of radar observation is 45°, the polarization mode is HH polarization.

Setting the plane parameters ${{\varepsilon }_{r}}=25$, $h=0.005m$, $l=0.01m$, and the cube target parameters ${{\varepsilon }_{r}}=75$, $h=0.002m$, $l=0.001m$, so the reference SAR image can be obtained. According to the obtained SAR images perform differentiable inversion optimization. Based on the initial parameters of the target minimize the loss function of Eq. (\ref{deqn_ex38}), and after multiple iterations, the optimized SAR image can be obtained. Columns (c), (d) and (e) are SAR images at azimuth angles of 0°, 120° and 240°. (b) is the loss function during the optimization process. The optimized cube parameter is ${{\varepsilon }_{r}}=73.95$, $h=0.00209m$, $l=0.00104m$, as shown in Table \ref{tab3}.

\begin{figure}[t]
\centering
\includegraphics[width=3.5in]{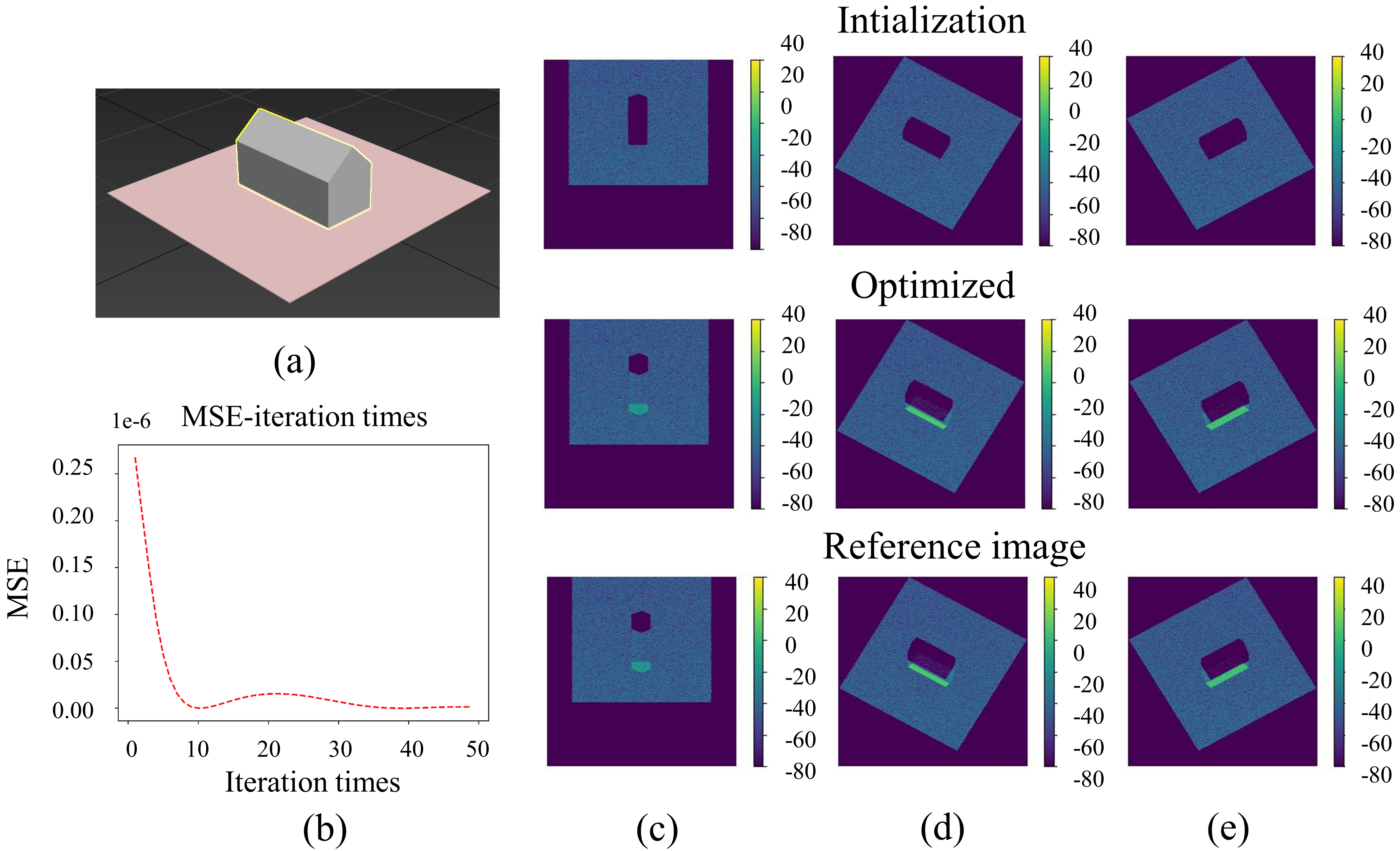}
\caption{SAR image simulation and differentiable optimization results in simplified building scenarios. (a) (b) (c) (d) (e) has the same meaning as in Fig \ref{fig_9}.}
\label{fig_10}
\end{figure}

\begin{table}[t]
\caption{Simplified building material parameter optimization and comparison with Ground Truth\label{tab:table4}}
\centering
\begin{tabular}{>{\rule[0ex]{0pt}{2ex}}c c c c}
\Xhline{1pt}  
Parameters & Initialization & Optimized & Ground Truth \\
\hline
${{\varepsilon }_{r}}$ & 1 & 7.42 & 6.885\\
\hline
$h$(m) & 0.0001 & 0.019 & 0.02\\
\hline
$l$(m) & 0.0001 & 0.009 & 0.01\\
\Xhline{1pt}  
\end{tabular}
\label{tab4}
\end{table}

As shown in Fig. \ref{fig_10}, scene consists of a plane and a simplified building. Setting the building parameters ${{\varepsilon }_{r}}=6.885$, $h=0.02m$, $l=0.01m$. In the same way, the optimization results of the building can be obtained after differentiable iterations. The optimized parameters is ${{\varepsilon }_{r}}=7.42$, $h=0.019m$, $l=0.009m$, as shown in Table \ref{tab4}.

Through experiments in two scenarios, the differentiability of the SPM-based ray tracing simulation model is verified, and more realistic SAR images can be obtained through automatic differentiation optimization of input parameters.

\subsection{Learning CSVBSDF parameters based on measured SAR images}
In this subsection, the surface scattering CSVBSDF parameters are learned from measured SAR images. SAR measured data is obtained by flying an antenna-equipped drone around a building. Table \ref{tab5} shows the system parameters of the detection radar.

\begin{figure}[t]
\centering
\includegraphics[width=3.5in]{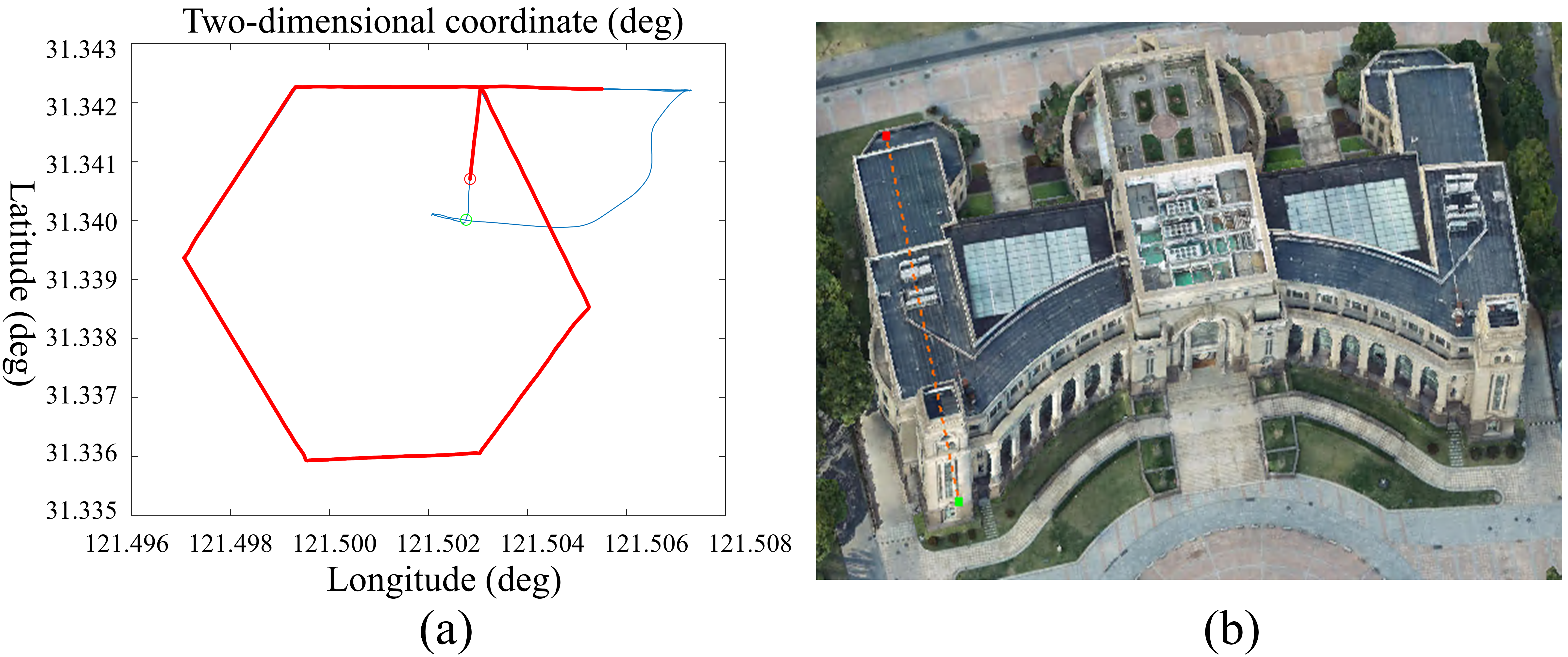}
\caption{SAR detection trajectories and optical images of complex buildings.}
\label{fig_11}
\end{figure}

\begin{table}[htbp]
\caption{Radar system parameters\label{tab:table5}}
\centering
\begin{tabular}{>{\rule[0ex]{0pt}{2ex}} >{\centering\arraybackslash}m{3cm} 
>{\centering\arraybackslash}m{5cm}}
\Xhline{1pt}  
Parameters & Value \\
\hline
Waveform & FMCW \\
\hline
Scanning method & Strip type, front and right side view \\
\hline
Center frequency & 15.2GHz \\
\hline
Platform relative height & 170m--475m \\
\hline
Pulse repetition frequency & 2000Hz \\
\hline
Azimuth beamwidth & 6° \\
\hline
Range Beamwidth & 20° \\
\hline
Platform speed & 10m/s \\
\hline
Down view & 60° \\
\hline
Sampling frequency & 25MHz \\
\hline
Bandwidth & 1200MHz \\
\hline
System power & 5W \\
\hline
Scene range & 173.60m*178.79m \\
\Xhline{1pt}  
\end{tabular}
\label{tab5}
\end{table}

\begin{figure*}[t!]
\centering
\includegraphics[width=7in]{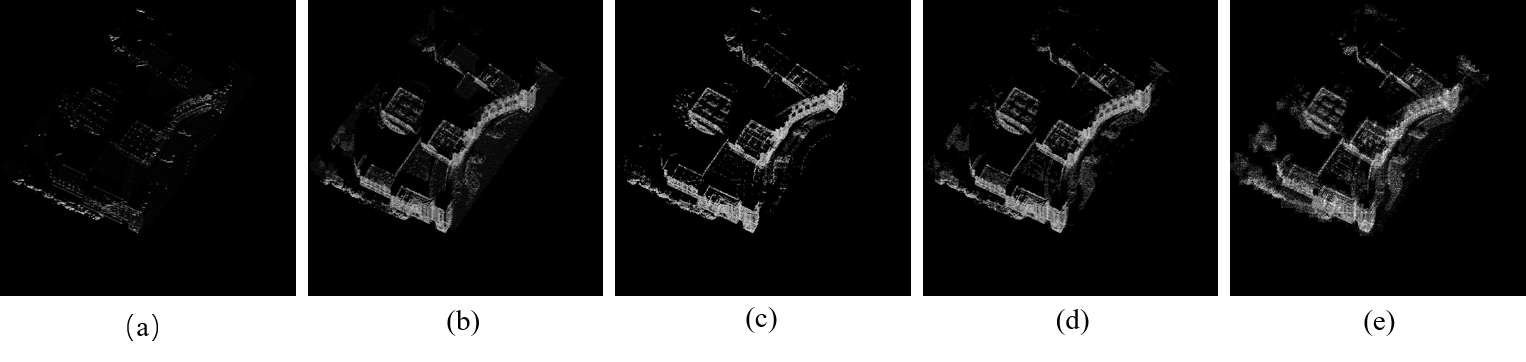}
\caption{Schematic diagram of SPM, KA and double-scale model learning results. (a) is the initial SAR image simulated with an initial value; (b) is the result of the SPM scattering model; (c) is the result of the KA scattering model; (d) is KA+SPM double-scale scattering model results; (e) is the ground truth measured SAR image at 0° azimuth angle after removing the background.}
\label{fig_12}
\end{figure*}

We set the same observation parameters as the actual measured parameters for simulation, and the architectural geometry was modeled using 3D oblique photography technology. Random parameter $\boldsymbol{\zeta} =(h_0,l_0,{{{\varepsilon }_{r}}_0})$ is initially set, and the parameter $\boldsymbol{\zeta}$ is iteratively optimized and learned based on the difference between the measured and simulated images. Firstly, learning based on one SAR image as shown in Fig. \ref{fig_12}. It can be observed that the simulation results obtained by the double-scale scattering model are closer to the ground truth.

\begin{figure*}[t!]
\centering
\includegraphics[width=7in]{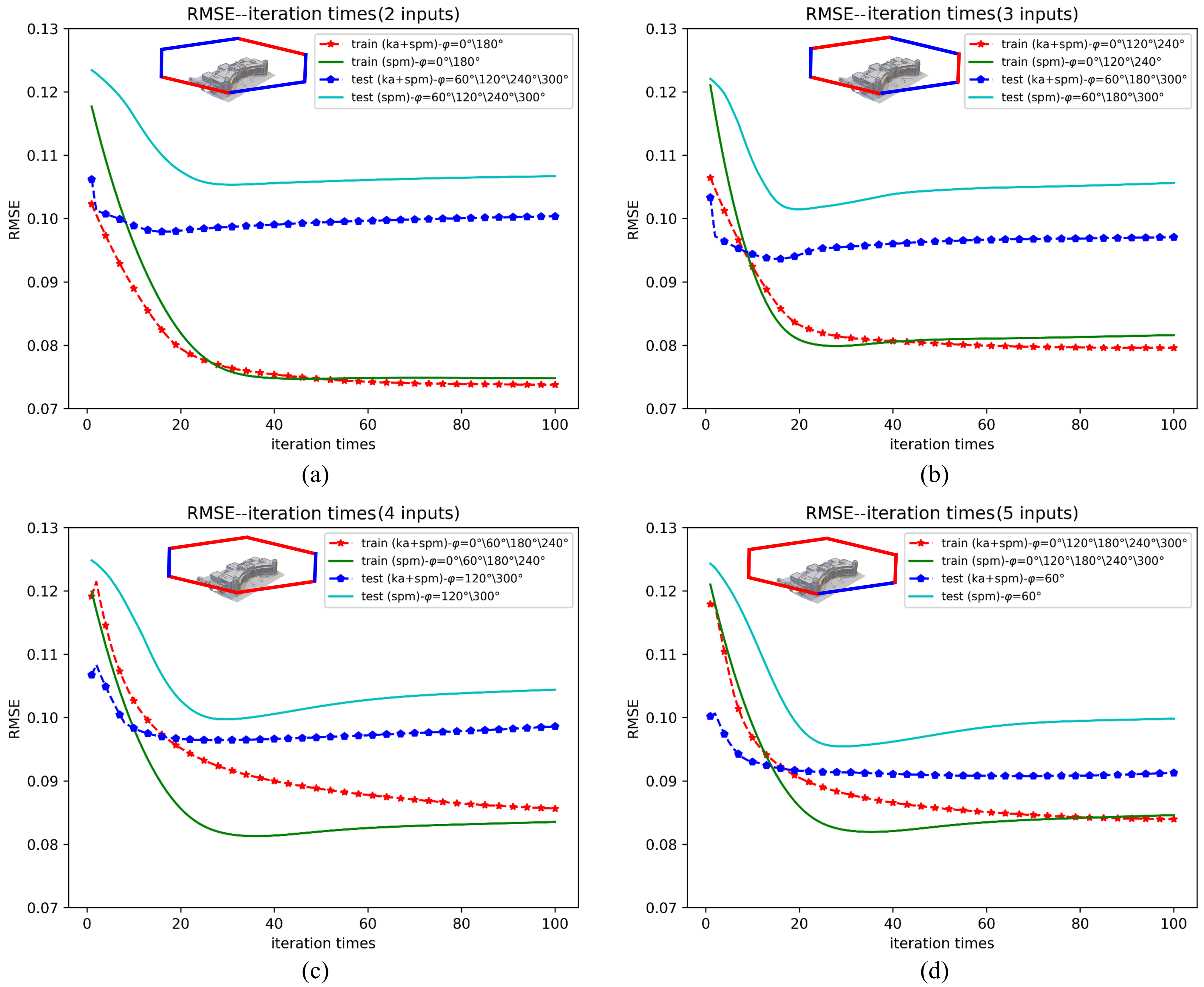}
\caption{Changes in the RMSE curves of different viewing angles participating in the optimization, including the viewing angles participating in the optimization and the viewing angles not participating in the optimization, using SPM and KA+SPM double-scale scattering models for experiments. (a) (b) (c) (d) represent the curves of training and testing RMSE and the number of iterations for 2, 3, 4, and 5 inputs reference number images respectively.}
\label{fig_13}
\end{figure*}

\subsubsection{Experimental setup}
Differentiable learning experiments require measured SAR images at different azimuth angles for training and testing. The flight trajectory of the Unmanned Aerial Vehicle (UAV) starts from the red circle in Fig. \ref{fig_11}(a) and circles clockwise. The flight azimuth angles are: 0°, 60°, 120°, 180°, 240°, 300°. The radar raw echoes are captured by the antenna and imaged from different azimuth angles, producing a SAR image on each side of the hexagonal trajectory. Fig. \ref{fig_11}(b) is an optical image of a complex building, and the imaging result is shown in Fig. \ref{fig_15}(a).

\subsubsection{Experimental results and analysis}

\begin{figure*}[htbp]
\centering
\includegraphics[width=7in]{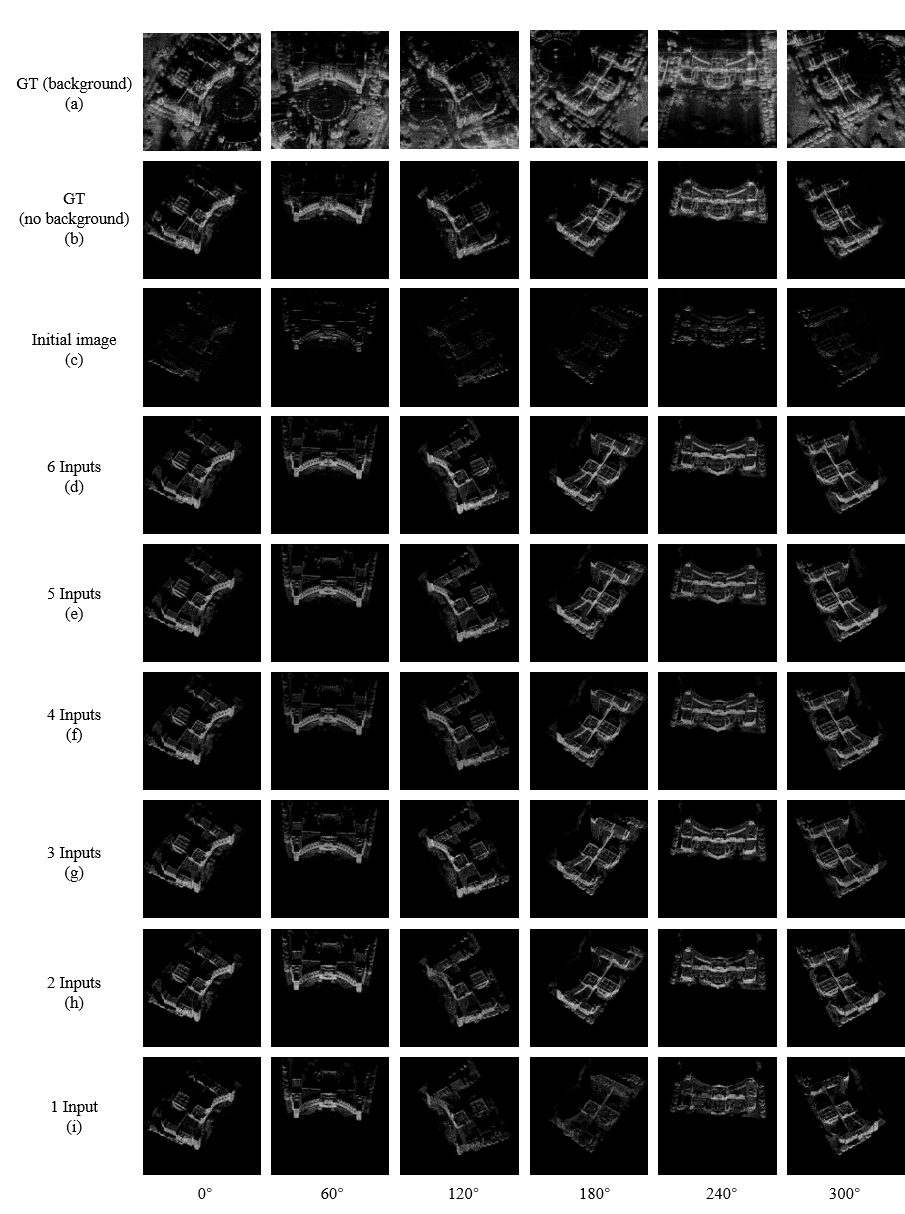}
\caption{Double-scale scattering model SAR inversion optimization experiment with different viewing angles participating.}
\label{fig_15}
\end{figure*}

We conduct inverse gradient optimization experiments on SAR measured images with different numbers of viewing angles, as shown in Fig. \ref{fig_13}. RMSE is the root mean square error between the simulation results and the measured image. (a) is 0°, 180° 2 inputs participate in training optimization, and the curves of RMSE and iterations are tested for images of 60°, 120°, 240°, and 300°; (b) is 0°, 120°, 240° 3 inputs participate in training optimization, 60°, 180°, 300° images are tested RMSE and iteration number curves; (c) is 0°, 60°, 180°, 240° 4 inputs participate in training optimization, 120°, 120°, The curve of the RMSE and the number of iterations of the 300° image for testing; (d) is 5 inputs of 0°, 120°, 180°, 240°, and 300° participate in training optimization, and the RMSE of the 60° image for testing and the number of iterations curve. The green and cyan solid lines represent the SPM scattering model, and the red and blue symbolic lines represent the double-scale scattering model. The optimized SAR images correspond to (e) (f) (g) (h) in Fig. \ref{fig_15} respectively.

It can be observed from Fig. \ref{fig_13} that as the number of perspectives participating in training increases, the RMSE convergence values of the two scattering model tests become smaller. And the difference between the convergence RMSE of the training image and the convergence RMSE of the test image is getting smaller and smaller, indicating that with the increase of SAR images from different viewing angles participating in the training, the parameter optimization results will be better and the generalization ability will be better. On the other hand, under any number of training views, the RMSE of the KA+SPM double-scale scattering model under the test image is significantly smaller than the SPM single-scale scattering model. The optimization results of the double-scale model are better than those of the SPM model.

\begin{figure}[t!]
\centering
\includegraphics[width=3.5in]{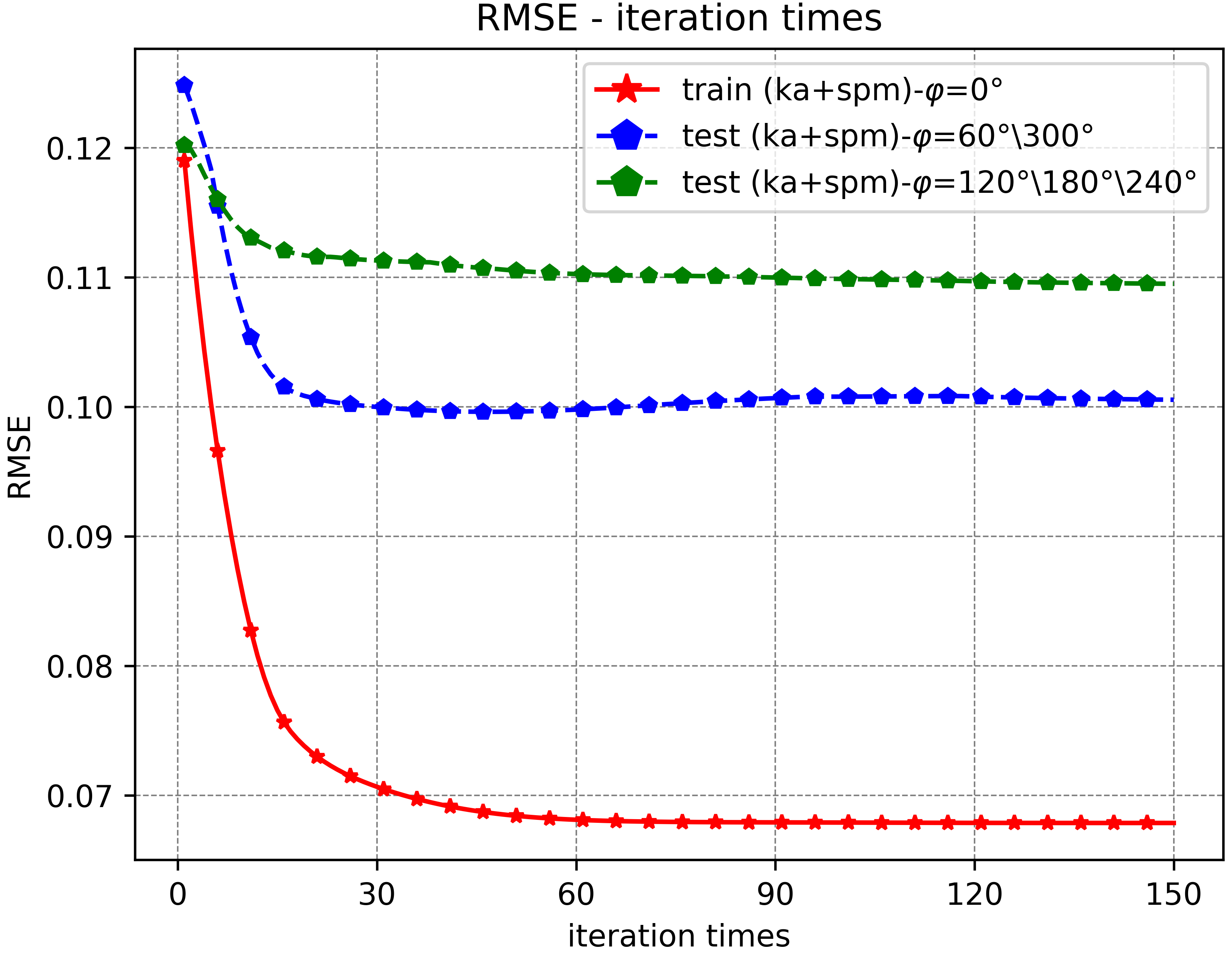}
\caption{Only one true SAR image with a viewing angle of 0° participates in the optimization training, and as the number of iterations increases, the RMSE with the true value at other viewing angles is calculated. According to whether  they are adjacent, they are categorized into adjacent angles of 60° and 300°, and relative angles of 120°, 180°, and 240°.}
\label{fig_14}
\end{figure}

We consider 0° as the central viewing angle, where 60° and 300° represent adjacent angles, and 120°, 180°, and 240° represent relative angles. As shown in Fig. \ref{fig_14}, the converged RMSE for 0° viewing angle is the lowest value involved in training. The converged RMSE for adjacent angles at 60° and 300° is lower than relative angles at 120°, 180°, and 240°. This is because the surface elements illuminated by the radar at adjacent viewing angles of 60° and 300° overlap more with the surface elements illuminated by the radar at the 0° viewing angle. This can prove that the visibility of the surface elements has a positive impact on the optimization results. Each surface element is irradiated at least once to ensure that the material parameters of the surface element can be optimized.

Fig. \ref{fig_15} shows the double-scale scattering model SAR inversion optimization experiment with different viewing angles participating: (a) shows the measured SAR images, and (b) shows the measured SAR images after removing the background. (c) is the simulated SAR images given a set of initial values $\boldsymbol{\zeta} =(h_0,l_0,{{{\varepsilon }_{r}}_0})$; (d) exhibits the SAR images after optimization with 6 inputs participating in parameter inversion; (e) illustrates the images after optimization with 5 inputs participating in inversion at 0°, 120°, 180°, 240°, and 300°, with 60° being the simulated viewing angle based on optimized parameters; (f) shows the optimized images with 4 inputs participating at 0°, 60°, 180°, and 240°; (g) demonstrates optimization with 3 inputs at 0°, 120°, and 240°; (h) displays the SAR images after optimization with 2 viewing angles at 0° and 180°; (i) depicts the SAR images after optimization with only 1 viewing angle at 0°, with the rest being simulated images based on parameters optimized for the 0° viewing angle.

\begin{figure*}[htbp]
\centering
\includegraphics[width=7in]{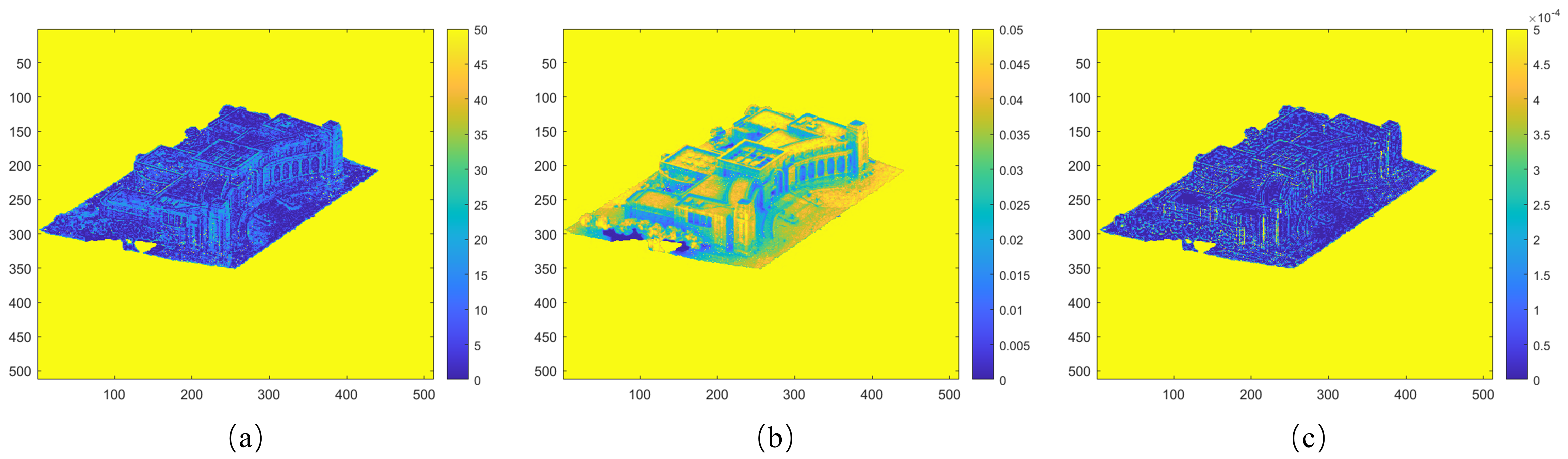}
\caption{3D diagram of learned CVBSDF parameters. (a) is the learned spatially varying ${{\varepsilon }_{r}}$; (b) is the learned microsurface spatially varying $l$; (c) is the learned spatially varying $h$ of the microsurface roughness.}
\label{fig_16}
\end{figure*}

Fig. \ref{fig_16} shows a 3D diagram of the CSVBSDF parameters after DRT differentiable learning. The 3D building surface scattering parameters present spatially varying distribution characteristics, and different parameters are within reasonable intervals.

The complex structure of the building covers many different media, such as marble, concrete, plastic, glass, asphalt, metal, etc., and it is surrounded by surface types such as soil, grass, trees, etc. Therefore, according to the traditional simulation method, a large number of scattering-related parameters need to be manually set for simulation. Manually setting parameters is usually based on experience and intuition, which is which is labor-intensive and may not be accurate. Slightly deviated parameter values may lead to inaccurate or unreliable results. Secondly, there may be complex interactions and couplings between parameters, and manual setting of parameters may not fully consider these complexities. Manually setting parameters may also require a lot of trial and error and adjustment, which consumes time and resources. Even so, it is often difficult to find the optimal parameters. Therefore, optimizing the CSVBSDF parameters at any position can greatly improve the simulation accuracy and efficiency.

\begin{table}[htbp]
\caption{Effect of sampling number on optimization results\label{tab:table6}}
\centering
\begin{tabular}{>{\rule[0ex]{0pt}{2ex}}c c}
\Xhline{1pt}  
Samples Per Unit Area (SPUA) & RMSE After Convergence \\
\hline
8 & 0.01696 \\
\hline
16 & 0.01537\\
\hline
32 & 0.01399\\
\hline
64 & 0.00907\\
\hline
\textbf{128} & \textbf{0.00556}\\
\hline
256 & 0.00801\\
\Xhline{1pt}  
\end{tabular}
\label{tab6}
\end{table}

\begin{table}[htbp]
\caption{Time consumption of forward simulation and reverse optimization\label{tab:table7}}
\centering
\begin{tabular}{>{\rule[0ex]{0pt}{2ex}} >{\centering\arraybackslash}m{1.2cm} 
>{\centering\arraybackslash}m{1.7cm} >{\centering\arraybackslash}m{0.7cm} 
>{\centering\arraybackslash}m{1.5cm} >{\centering\arraybackslash}m{1.5cm}}
\Xhline{1pt}  
Number of Facets & Resolution(m) & SPUA & Forward Simulation Time(s) & Reverse Optimization Time(s) \\
\hline
449899 & (0.41, 0.47) & 32 & 1.6920 & 0.9692\\
\hline
449899 & (0.41, 0.47) & 64 & 2.0341 & 1.1274\\
\hline
449899 & (0.41, 0.47) & 128 & 2.8441 & 1.4096\\
\hline
900391 & (0.41, 0.47) & 32 & 1.7070 & 0.9906\\
\hline
900391 & (0.41, 0.47) & 64 & 2.1240 & 1.1400\\
\hline
900391 & (0.41, 0.47) & 128 & 2.8634 & 1.4950\\
\hline
900391 & (0.82, 0.94) & 128 & 1.0797 & 0.6802\\
\hline
900391 & (1.64, 1.88) & 128 & 0.5253 & 0.5317\\
\Xhline{1pt}  
\end{tabular}
\label{tab7}
\end{table}

\subsubsection{DRT algorithm performance analysis}
Experiments on the influence of the sampling number on the optimization results for a single view of the building show that the radar transmitter's ray samples per unit area (SPUA) converges to the minimum RMSE when the number of samples is 128. As shown in Table \ref{tab6}, this does not necessarily imply that a larger SPUA will lead to better optimization results.

Table \ref{tab7} shows the forward simulation and reverse optimization time for the building and its surrounding scene in Fig. \ref{fig_11}. The scene size is 173.60m * 178.79m. It is demonstrated that the resolution has a significant impact on the speed of forward simulation and reverse optimization. The number of facets has a minimal effect on speed because the GPU accelerates ray tracing for intersections in parallel. For targets with about 900,000 facets, both forward and reverse optimization times are approximately a few seconds.

\section{Conclusion}

To obtain physically accurate gradient estimates, we develop a novel GPU-based Monte Carlo differentiable ray tracing SAR imaging simulation method that provides unbiased gradient estimates for CSVBSDF surface scattering parameters. To further enhance robustness, the algorithm adopts an end-to-end learning strategy that utilizes geometric targets and physically based scattering rendering models to generate physically reasonable surface scattering parameters. Using forward simulated SAR images and reference measured SAR images, we demonstrate that our method can achieve high-quality CSVBSDF parameter learning. Through multiple comparative verification experiments, DRT confirmed the impact of surface visibility on optimization results, in which ensuring that the surface is illuminated at least once is effective for surface scattering parameter learning. Even for highly complex objects or scenes, DRT requires only a few observation views to learn the surface scattering CSVBSDF parameters. Imaging with learned parameters can produce highly realistic SAR images while maintaining the algorithm's differentiable learning speed.

Although based on the measured images, this paper realizes the differentiable surface scattering parameters learning of ray tracing based on the physical mechanism, which provides a new idea for SAR simulation. However, there are still some improvements that need to be researched in the future:

1) Geometry optimization and reconstruction. The differentiable ray tracing parameter learning method proposed in this article only learns surface scattering parameters and does not optimize the 3D geometric structure. For some geometric models with inaccurate target modeling, geometric optimization based on measured SAR images will obtain more realistic simulated SAR images.

2) Microfacet scattering model learning. The current microfacet scattering model is not optimal. In the future, the fitting ability of network operators can be used to optimize the scattering model by performing equivalent network mapping with differentiable operators. Suppose the scattering model is integrated into a neural network with strong learning ability. In this case, it also addresses the lack of physics in neural networks.

\bibliographystyle{IEEEtran}
\bibliography{ref}

\vfill

\end{document}